\documentclass{article}
\usepackage{iclr2027_conference,times}

\usepackage{amsmath,amsfonts,bm}

\def\eqref#1{equation~\ref{#1}}
\def\1{\bm{1}}

\DeclareMathAlphabet{\mathsfit}{\encodingdefault}{\sfdefault}{m}{sl}
\SetMathAlphabet{\mathsfit}{bold}{\encodingdefault}{\sfdefault}{bx}{n}

\usepackage{booktabs}
\usepackage{multirow}
\usepackage{amssymb}
\usepackage{pifont}
\usepackage{fontawesome5}
\usepackage{graphicx}
\usepackage{wrapfig}
\usepackage{needspace}
\usepackage{tabularx}
\usepackage{longtable}
\usepackage{array}
\usepackage{enumitem}
\usepackage{xcolor}
\usepackage{hyperref}
\hypersetup{hidelinks}
\usepackage{url}

\title{Towards Robust Long-Lived Agents: Benchmarking Experience Lifecycles in Evolving Environments}

\author{Shijue Huang$^{1,4}$\thanks{Equal contribution.} \quad Hangyu Guo$^1$\footnotemark[1] \quad Hengheng Zhang$^3$ \quad Guo An$^3$ \quad Zhitao He$^1$ \\
\textbf{Qi Tian$^{2,3}$\thanks{Corresponding authors.} \quad Yiren Feng$^{1,4}$\footnotemark[2]} \\
\textnormal{$^1$The Hong Kong University of Science and Technology} \\
\textnormal{$^2$Guangdong Laboratory of Artificial Intelligence and Digital Economy (SZ)} \\
\textnormal{$^3$Huawei Inc. \quad $^4$HKUST Shenzhen Research Institute} \\[4pt]
{\small\normalfont
\href{https://ReScopeBench.github.io/}{\faHome\enspace\texttt{ReScopeBench.github.io}}
\qquad
\href{https://github.com/JoeYing1019/ReScopeBench}{\faGithub\enspace\texttt{github.com/JoeYing1019/ReScopeBench}}}
}
\iclrfinalcopy

\newcommand{\resc}{ReScopeBench}

\definecolor{markgreen}{HTML}{267A3F}
\definecolor{markred}{HTML}{A33A3A}
\newcommand{\cmark}{\textcolor{markgreen}{\ding{51}}}
\newcommand{\xmark}{\textcolor{markred}{\ding{55}}}


\begin{document}
\maketitle
\lhead{Preprint}

\begin{abstract}
As agents become more capable, they are moving beyond one-off tasks toward sustained assistance across successive user goals. Such assistance requires agents to discover how unfamiliar environments work, reuse what they learn across tasks, and keep that experience current as environments evolve. Existing benchmarks increasingly study experience reuse and adaptation, but provide limited insight into which parts of accumulated experience agents revise, retain, or lose. To address this gap, we introduce \resc{}, a benchmark for the experience lifecycle across 18 interactive environments. It distinguishes structural experience about how an environment is organized and accessed from rule experience about reusable principles and constraints that guide decisions across tasks. To study how experience is acquired, reused, and revised, we propose an experience-first construction approach. It builds task streams around shared experience and introduces silent environmental changes. We evaluate task performance and track changes in saved experience throughout each stream. Across 20 models, we find that retained experience improves task performance, but agents struggle to keep it current as environments change. Even when the required experience is correctly recorded, agents do not always fulfill requests successfully. These findings identify selective maintenance and effective use as complementary challenges for long-lived agents.
\end{abstract}

\raggedbottom
\section{Introduction}
Recent advances have strengthened agents' ability to complete complex tasks in interactive environments \citep{openai2026gpt6astrasystemcard,anthropic2026claudeopus55systemcard,DBLP:journals/corr/abs-2509-02544,DBLP:journals/tmlr/GonzalezPumariegaTLYLW26,DBLP:journals/corr/abs-2607-23588}. This progress opens the way to sustained assistance across successive user goals. In an initially unfamiliar environment, such assistance requires the agent to discover how resources are organized and which rules guide its actions. When new requests depend on the same resources and rules, agents can reuse earlier discoveries to guide their work. Continued reuse also requires keeping this experience current when the environment changes.

% These requirements make learning from interaction central to sustained assistance. Self-improving agents pursue this goal by using execution and feedback to improve later problem solving \citep{DBLP:conf/iclr/ZhangHLLC26,DBLP:journals/corr/abs-2603-19461,DBLP:journals/corr/abs-2606-26294}. Benchmarks complement this work by assessing sustained task execution \citep{DBLP:journals/corr/abs-2605-30434,liu2026skillhorizonconstructinglonghorizonagent} and the use of experience across tasks. Memory benchmarks evaluate retrieval and updating of information from supplied histories \citep{DBLP:conf/acl/MaharanaLTBBF24,DBLP:conf/iclr/WuWYZCY25,DBLP:conf/acl/Tan000DD25,DBLP:journals/corr/abs-2507-05257}. Interactive benchmarks extend this scope to experience that agents acquire through their own actions and reuse on later tasks \citep{DBLP:journals/corr/abs-2505-11942,DBLP:journals/corr/abs-2602-16313,DBLP:journals/corr/abs-2604-17308,DBLP:journals/corr/abs-2605-24117}. More recent benchmarks, including AgingBench and EvoArena, examine memory maintenance and adaptation as information and environments evolve \citep{DBLP:journals/corr/abs-2605-26302,DBLP:journals/corr/abs-2606-13681}.

To evaluate these abilities, existing benchmarks examine experience reuse across tasks and adaptation to change. Memory benchmarks test retrieval and reasoning over supplied histories, with some also assessing information updates \citep{DBLP:conf/acl/MaharanaLTBBF24,DBLP:conf/iclr/WuWYZCY25,DBLP:conf/acl/Tan000DD25,DBLP:journals/corr/abs-2507-05257}. Interactive and skill-oriented benchmarks assess experience acquired through agents' own actions and reused or refined in later tasks \citep{DBLP:journals/corr/abs-2505-11942,DBLP:journals/corr/abs-2602-16313,DBLP:journals/corr/abs-2604-17308,DBLP:journals/corr/abs-2605-24117}. AgingBench and EvoArena further examine memory maintenance and adaptation as information and operating conditions change \citep{DBLP:journals/corr/abs-2605-26302,DBLP:journals/corr/abs-2606-13681}.

However, existing evaluations do not fully show whether agents update outdated experience while keeping the parts that remain valid after the same environmental change. Task success alone cannot answer this question. For example, a delivery directory may move while its delivery policy stays the same. An agent may find the new directory and complete the task but leave the old path in memory. It may also rewrite the saved entry and lose policy details that are still valid. These cases can receive the same task score as an update that corrects the path and keeps the policy. When queries and task actions share a limited budget, repeated lookups leave fewer resources for completing requests. We therefore evaluate task results together with the experience agents save. We track what agents learn, update, retain, or lose across tasks and environmental changes. Table~\ref{tab:intro-comparison} compares existing benchmarks and \resc{} along these dimensions.

\begin{table}[!t]
\centering
\fontsize{8}{10}\selectfont
\setlength{\tabcolsep}{3pt}
\renewcommand{\arraystretch}{1.15}
% Vertically center all cells, including single- and two-line headers.
\renewcommand{\tabularxcolumn}[1]{m{#1}}
\caption{Comparison of benchmarks for agent memory and experience. MAB and LAB denote MemoryAgentBench and LifelongAgentBench, respectively.}
\label{tab:intro-comparison}
% Coverage concerns documented benchmark protocols, including published diagnostic analyses.
% Budgeted exploration denotes domain-specific information-gathering costs, not generic token or turn limits.
% Revision and retention concern direct assessment of saved content, not task-level repair alone.
\begin{tabularx}{\linewidth}{@{}>{\raggedright\arraybackslash}m{0.28\linewidth}>{\centering\arraybackslash}m{0.12\linewidth}>{\hsize=.77\hsize\linewidth=\hsize\centering\arraybackslash}X>{\hsize=1.15\hsize\linewidth=\hsize\centering\arraybackslash}X>{\hsize=.90\hsize\linewidth=\hsize\centering\arraybackslash}X>{\hsize=1.03\hsize\linewidth=\hsize\centering\arraybackslash}X>{\hsize=1.15\hsize\linewidth=\hsize\centering\arraybackslash}X@{}}
\toprule
 & \multicolumn{3}{c}{\textbf{Experience}} & \multicolumn{2}{c}{\textbf{Environmental Change}} & \textbf{Assessment} \\
\cmidrule(lr){2-4}\cmidrule(lr){5-6}\cmidrule(l){7-7}
Benchmark & Acquisition & Reuse & Budgeted\newline Exploration & Silent & Multiple & Revision \&\newline Retention \\
\midrule
LoCoMo~\citep{DBLP:conf/acl/MaharanaLTBBF24} & Provided & \xmark & \xmark & \xmark & \xmark & \xmark \\
LongMemEval~\citep{DBLP:conf/iclr/WuWYZCY25} & Provided & \xmark & \xmark & \cmark & \xmark & \xmark \\
MemBench~\citep{DBLP:conf/acl/Tan000DD25} & Provided & \xmark & \xmark & \xmark & \xmark & \xmark \\
MAB~\citep{DBLP:journals/corr/abs-2507-05257} & Provided & \cmark & \xmark & \cmark & \cmark & \xmark \\
LAB~\citep{DBLP:journals/corr/abs-2505-11942} & Interactive & \cmark & \xmark & \xmark & \xmark & \xmark \\
MemoryArena~\citep{DBLP:journals/corr/abs-2602-16313} & Interactive & \cmark & \xmark & \xmark & \xmark & \xmark \\
SkillFlow~\citep{DBLP:journals/corr/abs-2604-17308} & Interactive & \cmark & \xmark & \xmark & \xmark & \xmark \\
SkillEvolBench~\citep{DBLP:journals/corr/abs-2605-24117} & Interactive & \cmark & \xmark & \xmark & \xmark & \xmark \\
AgingBench~\citep{DBLP:journals/corr/abs-2605-26302} & Mixed & \cmark & \xmark & \xmark & \cmark & \xmark \\
EvoArena~\citep{DBLP:journals/corr/abs-2606-13681} & Mixed & \cmark & \xmark & \cmark & \cmark & \xmark \\
\midrule
\textbf{\resc{}} & Interactive & \cmark & \cmark & \cmark & \cmark & \cmark \\
\bottomrule
\end{tabularx}
\end{table}

To support this evaluation, we introduce \resc{}, a benchmark spanning 18 initially unfamiliar environments. We focus on two complementary types of experience that agents can acquire in these environments. \emph{Structural experience} describes how resources are organized, related, and accessed. \emph{Rule experience} captures reusable principles and constraints that guide decisions across tasks within an environment. To study how the same experience is acquired, reused, and revised, we propose an \emph{experience-first} construction approach. We first identify structural and rule experience that agents can discover through interaction, then build varied tasks in which agents can reuse it. Local changes invalidate selected parts of this experience while leaving others valid. Changes are unannounced, so agents must detect them through interaction. To examine experience maintenance at different stages of interaction and beyond a single change, we vary change positions and include multiple changes in some streams. Without exploration costs, agents could repeatedly rediscover information instead of maintaining useful experience. We therefore place discovery and execution under environment-specific resource budgets, encouraging agents to reuse valid experience and selectively recheck what may have changed. We evaluate task performance with and without retained experience and assess saved content against private references that reflect the current environment.

We evaluate 20 commercial and compact open-weight models on \resc{} and find that retaining experience improves average task performance. Agents generally record structural experience more completely than rule experience. When environments change, they preserve much of the valid content but leave some outdated content uncorrected. More interaction after change does not always coincide with more explicit attempts to update saved experience. Even when the required experience is correctly recorded, agents can still fail to fulfill requests. These findings highlight the need to update experience without losing valid content and to use saved experience effectively to complete tasks.

Our contributions can be summarized as follows:
\begin{itemize}[leftmargin=1.4em,itemsep=0pt,topsep=1pt]
\item We introduce \resc{} to study the experience lifecycle in evolving environments. It evaluates task performance together with the acquisition, retention, and revision of saved structural and rule experience.
\item We construct 352 tasks across 18 interactive environments, organized into streams in which agents can discover and reuse structural and rule experience. The streams include 26 controlled environmental changes that make selected experience outdated while other experience remains valid.
\item Our evaluation of 20 models combines task outcomes, saved experience, and action traces. It reveals gaps in experience acquisition, incomplete revision of outdated content, and failures to fulfill requests even when the required experience is correctly recorded.
\end{itemize}

\flushbottom
\section{ReScopeBench}
\label{sec:method}

\subsection{Benchmark Overview}
\label{sec:overview}
\resc{} evaluates agents over task streams in initially unfamiliar interactive environments. We use $e$ to denote one environment, $T_e$ its stream length, and $t=1,\ldots,T_e$ the task position within that stream. For each task, agents receive a goal and descriptions of the available tools. They use these tools to discover resource relationships and operating rules, and can record these findings for reuse in later tasks.

Completing varied goals requires agents to locate relevant resources and understand how to act on them. We distinguish \emph{structural experience}, which describes how resources are organized, related, and accessed, from \emph{rule experience}, which captures reusable principles and constraints that guide decisions across tasks. For example, locating stored goods does not establish how they must be handled. Their location can change while handling rules remain valid, or vice versa. Separating these types lets us track their acquisition, revision after change, and retention of still-valid content.

To track how the same experience is acquired, reused, and revised across tasks, we propose an \emph{experience-first construction} approach. We first identify discoverable experience, then build varied tasks that reuse it and local changes that invalidate selected parts. Changes vary in timing and number and must be detected through interaction without explicit notice. Resource budgets limit exhaustive rediscovery, encouraging agents to reuse earlier findings. Figure~\ref{fig:pipeline} summarizes the construction process and lifecycle. We evaluate task performance with and without retained experience and assess saved content against references for the current environment.

\subsection{Benchmark Construction}
\label{sec:construction}
\begin{figure}[!t]
\centering
\includegraphics[width=\textwidth]{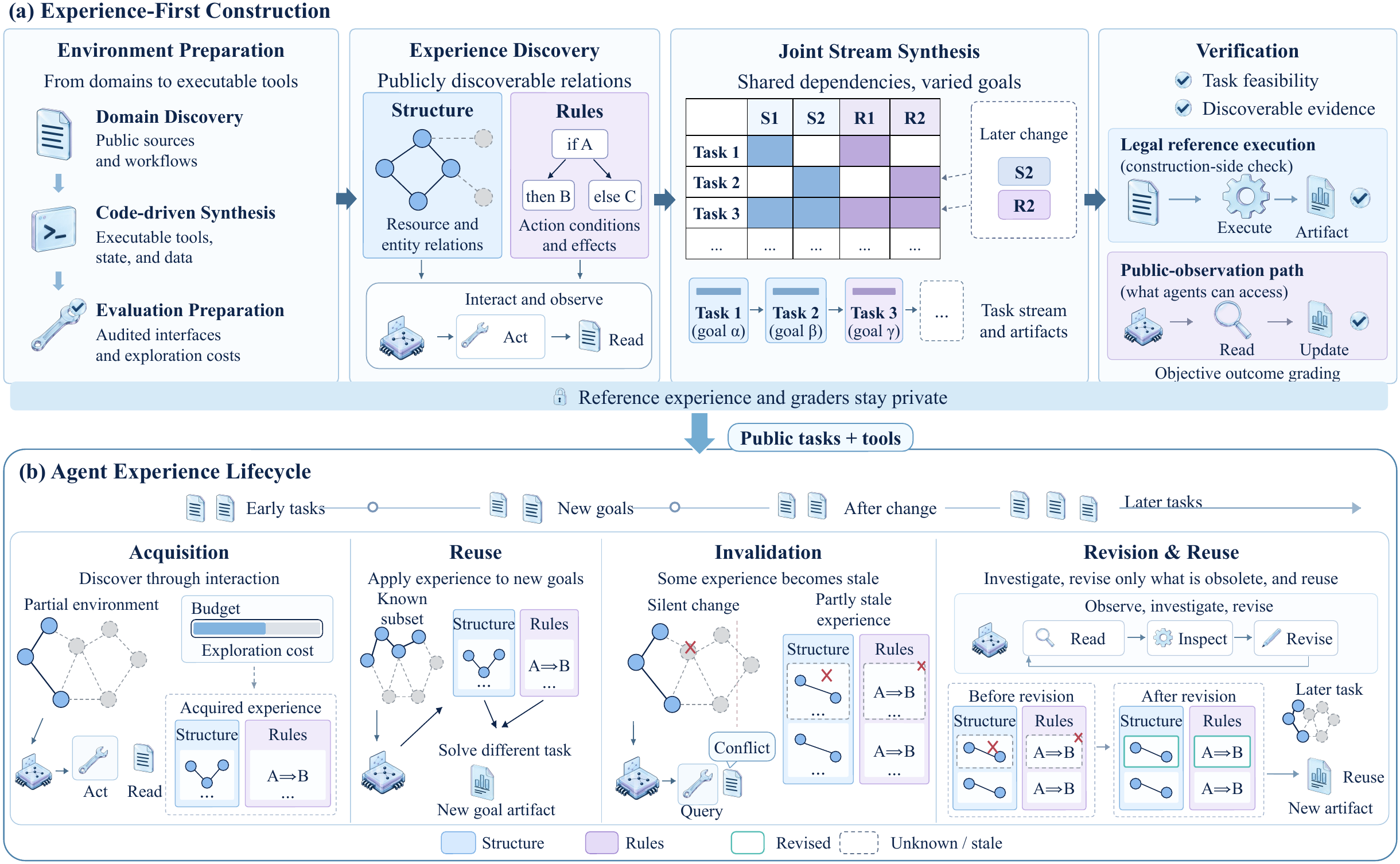}
\caption{ReScopeBench overview. (a) Experience-first construction connects executable environments, discoverable dependencies, and verified task streams. (b) Agents acquire, reuse, and selectively revise experience after silent changes.}
\label{fig:pipeline}
\end{figure}

As depicted in Figure~\ref{fig:pipeline}(a), construction proceeds through environment preparation, experience discovery, joint stream synthesis, and verification.

\paragraph{Environment Preparation.}
\label{sec:environment-synthesis}
We implement each environment $e$ as a simulated application in code, using application scenarios and public documentation to define its entities, workflows, and operations. Each simulation provides reproducible initial states and tools whose outputs reflect the current state. Tool descriptions explain how to use the interfaces, while local resource relationships and operating conditions must be discovered through interaction. Each environment specifies which tool operations incur costs and the limits on the resources they consume.

\paragraph{Experience Discovery.}
\label{sec:experience-synthesis}
The selected experience must support multiple task goals and be discoverable through the tools available to evaluated agents. We identify candidate experience by inspecting environment code and data, then verify its discoverability through tool interactions. For each selected item, we record its scope, the observations that reveal it, and the decisions and task outcomes that depend on it. These links form a private dependency graph for constructing tasks that reuse the experience and changes that invalidate selected parts.

To assess the saved content, we organize the selected experience within each environment $e$ into reference units $u$. Each unit groups related statements, called \emph{claims} $c$, about resource relationships or operating rules. Each claim has a scoring rubric specifying what the saved content must establish to count as correct. The references reflect the environment version at each task and remain private to the evaluator. Appendix~\ref{app:grove-lifecycle} illustrates a rubric and its application.

\paragraph{Joint Stream Synthesis.}
\label{sec:task-synthesis}
\label{sec:stream-synthesis}
To test reuse across different goals, we use the dependency graph to construct an ordered stream of $T_e$ tasks for environment $e$. The tasks share resource relationships and operating rules but vary in their requested objects and conditions.
\label{sec:environment-changes}
To test selective revision, we schedule local changes immediately before selected tasks. Their positions form the change schedule $\mathcal D_e$, with each $d\in\mathcal D_e$ denoting the first task affected by a change. Each change alters selected relationships or rules and updates their reference claims. For a given change and affected unit $u$, let $N_u$ denote the current-version claims describing information introduced or changed by that event. Later tasks use both changed and unchanged information. Changes are detectable through tool observations but are not announced in task instructions or boundary messages.

We complete the tasks in sequence using the tools available to agents, carrying state forward and applying scheduled changes. These tool-call sequences serve as private reference solutions. We use the resulting task artifacts and relevant final states to specify each task's delivery requirements. A private grader checks the agent's delivered results against these requirements and accepts equivalent valid outcomes. Appendix~\ref{app:grove-lifecycle} gives an example. We then compare the resource costs of reference solutions, local correction after changes, and exhaustive searches over candidate resources and rules. Budgets cover verified completion and correction costs while restricting exhaustive rediscovery.

\paragraph{Verification.}
\label{sec:verification}
Starting from each environment's initial state, we replay the reference solutions in task order and apply the scheduled changes. We check that each solution produces the required outputs within budget and that the grader assigns the expected score. We also test tool-call sequences that reveal updated information and support local correction within budget. Finally, we inspect task instructions and tool interfaces for leaked answers and unintended shortcuts. We freeze the tasks, tools, budgets, change schedules, and graders before evaluating models' ability to acquire, reuse, and revise experience.

\subsection{Benchmark Statistics}
\label{sec:statistics}

\begin{wrapfigure}[20]{R}{0.47\textwidth}
    \centering
    \includegraphics[width=\linewidth]{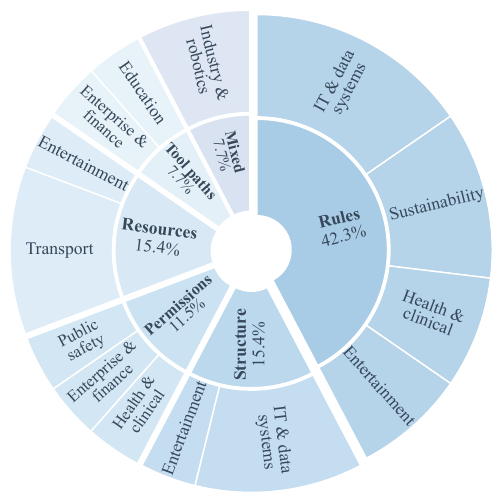}
    \caption{Distribution of change type and environment domain.}
    \label{fig:benchmark_statistics}
\end{wrapfigure}

ReScopeBench contains 352 tasks across 18 environments and nine application domains. Each environment exposes 3--18 tools (10.1 on average). Task streams have 16, 24, or 32 tasks and contain one to three environmental changes. The 26 changes occur at different positions, with the first change between tasks 5 and 13. Figure~\ref{fig:benchmark_statistics} groups these events by change type and environment domain.

We distinguish six mechanisms by what changes in the environment. \emph{Rule} changes alter the conditions governing valid actions and outcomes. \emph{Structure} changes alter resource locations, identifiers, or relationships. \emph{Permission} changes alter which actors are authorized to perform an operation. \emph{Resource} changes alter the availability of items, capacity, or time slots. \emph{Tool-path} changes alter which existing tools or tool sequences can complete an operation. \emph{Mixed} changes combine two or more mechanisms within the same event. These categories describe environmental changes, whereas structural and rule experience describe what agents learn. Appendix~\ref{app:environment-inventory} provides the environment inventory and change schedules.

\subsection{Evaluation Protocol and Metrics}
\label{sec:metrics}

\subsubsection{Evaluation Protocol}
For each model--environment pair, two streams start from the same initial state with identical tasks, tools, budgets, and change schedules. Agents are instructed to store structural experience in memory and rule experience in skills. The \emph{with-experience} condition retains these assets and retrievable history across tasks. The \emph{without-experience} condition clears them between tasks. Both conditions permit within-task learning and start each task with a fresh conversation. Environment state persists within each stream and can diverge as agents take different actions.

\subsubsection{Metrics}
We report three performance metrics for task outcomes and four experience metrics for the coverage, updating, and recovery of saved experience, each scored from 0 to 100 with higher values indicating better results. The experience metrics use post-task memory and skill snapshots from the with-experience stream. Appendix~\ref{app:metric-definitions} details aggregation.

\paragraph{Task performance.}
To assess whether retained experience helps agents complete tasks and how they perform when a change first affects a task, we report baseline performance (BP), experience-enabled performance (EP), and drift-onset performance (DP). For a fixed model, let $s^+_{e,t}$ and $s^-_{e,t}$ denote task scores with and without experience in environment $e$ at task $t$. We use the stream length $T_e$ and change schedule $\mathcal D_e$ from construction to define
% \begin{equation}
% \begin{aligned}
% \mathrm{BP}_e &= \frac{1}{T_e}\sum_{t=1}^{T_e}s^-_{e,t}, \\
% \mathrm{EP}_e &= \frac{1}{T_e}\sum_{t=1}^{T_e}s^+_{e,t}, \\
% \mathrm{DP}_e &= \frac{1}{|\mathcal D_e|}\sum_{d\in\mathcal D_e}s^+_{e,d}.
% \end{aligned}
% \end{equation}
\begin{equation}
\mathrm{BP}_e=\frac{1}{T_e}\sum_{t=1}^{T_e}s^-_{e,t},\quad \mathrm{EP}_e=\frac{1}{T_e}\sum_{t=1}^{T_e}s^+_{e,t},\quad \mathrm{DP}_e=\frac{1}{|\mathcal D_e|}\sum_{d\in\mathcal D_e}s^+_{e,d}.
\end{equation}
DP measures performance after executing the first affected task, including adaptation during that task.

\paragraph{Experience coverage.}
Task success does not show how much experience an agent has correctly saved for later tasks. We therefore use structural coverage (SC) and rule coverage (RC) to measure how much of the corresponding current-version reference inventories is correctly recorded. An LLM judge $\mathcal J$ receives the entire saved memory and skill corpus together with one reference experience unit. It assesses the unit's claims against their rubrics, regardless of where the content is stored. Each claim scores 1 only if all required and applicable checks pass, and 0 otherwise. SC and RC average claim scores within units, units within each type, and all task positions, then multiply by 100.

\paragraph{Experience updating.}
After a change, agents need to record the required updates while keeping experience that remains valid. We use update coverage (UC) to measure the fraction of update targets correctly recorded after the first affected task, and recovery timeliness (RT) to measure how early all update targets and previously correct, still-valid claims are recorded correctly together. For a fixed model and a change at position $d$ in environment $e$, consider each affected unit $u$ and its update targets $N_u$ defined during construction. Let $V_u$ contain the current-version counterparts of its still-valid claims credited after task $d-1$. The nonempty set $N_u$ is shared across models. Both sets remain fixed over an $L$-task window beginning with the first affected task and ending before the next change or at stream completion. Within this window, local position $k=1,\ldots,L$ corresponds to stream task $t=d+k-1$. Let $z_{c,k}$ indicate whether claim $c$ is correct after that task. We define
\begin{equation}
\mathrm{UC}_u=\frac{100}{|N_u|}\sum_{c\in N_u}z_{c,1},\qquad
\mathrm{RT}_u=\begin{cases}
100(L-\tau_u+1)/L, & \tau_u\text{ exists}, \\
0, & \text{otherwise}.
\end{cases}
\end{equation}
where $\tau_u=\min\{k:\forall c\in N_u\cup V_u,\ z_{c,k}=1\}$ is the first task at which all required claims are correct together. RT records this first completion, not whether correctness persists at later checkpoints. Because update targets do not depend on prior mastery, both metrics can include first acquisition as well as revision.

\section{Experiments}
\label{sec:experiments}

\subsection{Experimental Setup}
We evaluate 20 models from ten families: three compact open-weight models (Qwen3.5-9B, Qwen3.5-35B-A3B, and Qwen3.8-27B) \citep{DBLP:journals/corr/abs-2505-09388} and 17 models offered through commercial services, including models with public weights. The commercial group covers GPT \citep{openai2026gpt6astrasystemcard}, Claude \citep{anthropic2026claudeopus55systemcard}, DeepSeek \citep{DBLP:journals/corr/abs-2606-19348}, Doubao \citep{DBLP:journals/corr/abs-2504-13914}, Gemini \citep{DBLP:journals/corr/abs-2507-06261}, GLM \citep{DBLP:journals/corr/abs-2602-15763}, Kimi \citep{DBLP:journals/corr/abs-2607-24653}, MiMo \citep{DBLP:journals/corr/abs-2601-02780}, MiniMax \citep{DBLP:journals/corr/abs-2606-13392}, and Qwen3.8-Max. These reports describe the model families. Table~\ref{tab:main-results} lists the versions we evaluated.

All models use the Hermes agent harness \citep{nousresearch2026hermesagent} on the same 18 environments. Each model completes one with-experience stream and one without-experience stream per environment, following the protocol in Section~\ref{sec:metrics}.
Task outcomes are scored by deterministic graders. We use GPT-6 Astra \citep{openai2026gpt6astrasystemcard} to assess saved experience against the reference rubrics. Appendix~\ref{app:evaluation-details} provides inference settings, runtime limits, and detailed execution and scoring procedures.

\subsection{Main Results}
\label{sec:main-results}
Table~\ref{tab:main-results} reports the main results. We highlight three observations.

\noindent\textbf{(1) One compact model approaches the commercial median in immediate updating.}
Qwen3.8-27B records 29.05\% of the required updates after the first affected task, close to the commercial median of 29.75\%. Its task performance with experience shows a wider gap (62.48 versus 73.44). Both Qwen3.5 models score below 3 on each of these metrics.

\noindent\textbf{(2) Rule coverage remains uneven even among strong models.}
Structural coverage exceeds rule coverage for 17 of 20 models, with a mean gap of 4.79 points. This imbalance is not confined to models with low coverage. GPT-6 Astra and GPT-6.1 Sol have structural scores above 70, yet their rule scores trail by 9.51 and 10.97 points, respectively. Claude Opus 5.5 reverses this pattern: its rule coverage of 73.22 leads the table and exceeds its structural coverage of 71.70. Strong structural coverage therefore does not uniformly translate into equally complete recording of rules, and the balance between the two types remains model dependent.

\noindent\textbf{(3) High recovery scores can coexist with limited immediate updating.}
No model exceeds 37.63 in UC, leaving substantial gaps in recorded updates after the first affected task. Model rankings also differ between immediate updating and recovery timeliness. GPT-6 Astra ranks second in RT but tenth in UC, whereas Gemini-3.8-Flash ranks second in UC but ninth in RT. Strong recovery over subsequent tasks should therefore not be taken as evidence of a prompt initial response to change.

\begin{table}[!t]
\centering
\caption{Main results on \resc{}. BP: baseline performance; EP: experience-enabled performance; DP: drift-onset performance; SC: structural coverage; RC: rule coverage; UC: update coverage; RT: recovery timeliness. \textbf{Bold} and \underline{underlined} values denote the best and second-best distinct scores in each column, respectively; ties receive the same formatting.}
\label{tab:main-results}
\fontsize{9}{11}\selectfont
\setlength{\tabcolsep}{2.5pt}
\renewcommand{\arraystretch}{1.16}
\resizebox{\linewidth}{!}{%
\begin{tabular*}{1.06\linewidth}{@{\extracolsep{\fill}}ll*{8}{r}@{}}
\toprule
\multirow{2}{*}{\textbf{Provider}} & \multirow{2}{*}{\textbf{Model}} & \multicolumn{3}{c}{\textbf{Performance}} & \multicolumn{4}{c}{\textbf{Experience}} & \multirow{2}{*}{\textbf{Avg.}} \\
\cmidrule(lr){3-5}\cmidrule(lr){6-9}
& & BP & EP & DP & SC & RC & UC & RT & \\
\midrule
\multicolumn{10}{c}{\textit{Compact Open-Weight Models}} \\
\midrule
\raisebox{-0.15ex}{\includegraphics[width=1.1em,height=1.1em,keepaspectratio]{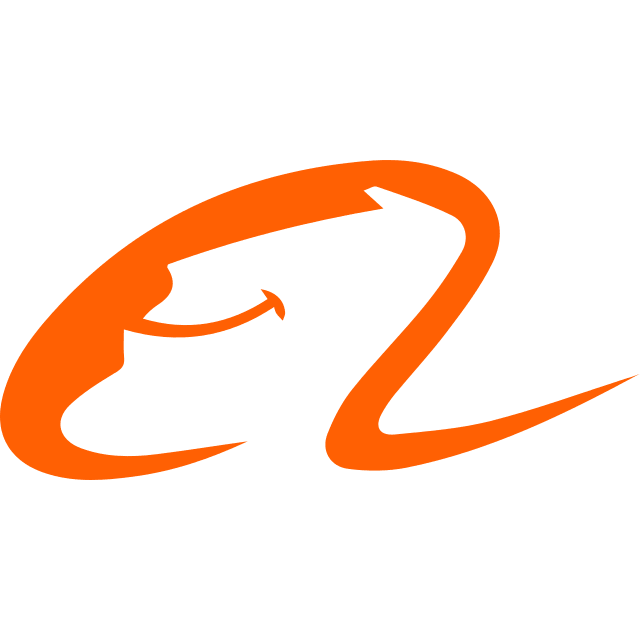}}\hspace{0.35em}Alibaba & \mbox{Qwen3.5-9B} & 0.51 & 0.89 & 0.74 & 15.14 & 18.43 & 1.67 & 2.95 & 5.76 \\
\raisebox{-0.15ex}{\includegraphics[width=1.1em,height=1.1em,keepaspectratio]{fig/provider_alibaba.png}}\hspace{0.35em}Alibaba & \mbox{Qwen3.5-35B-A3B} & 0.82 & 2.75 & 0.89 & 18.19 & 18.21 & 0.40 & 4.18 & 6.49 \\
\raisebox{-0.15ex}{\includegraphics[width=1.1em,height=1.1em,keepaspectratio]{fig/provider_alibaba.png}}\hspace{0.35em}Alibaba & \mbox{Qwen3.8-27B} & 20.07 & 62.48 & 38.21 & 60.90 & 55.18 & 29.05 & 59.17 & 46.44 \\
\midrule
\multicolumn{10}{c}{\textit{Commercial Models}} \\
\midrule
\raisebox{-0.15ex}{\includegraphics[width=1.1em,height=1.1em,keepaspectratio]{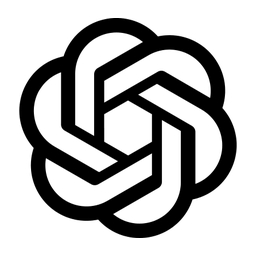}}\hspace{0.35em}OpenAI & \mbox{GPT-5.4} & 0.09 & 14.96 & 12.38 & 47.75 & 33.15 & 12.91 & 22.85 & 20.58 \\
\raisebox{-0.15ex}{\includegraphics[width=1.1em,height=1.1em,keepaspectratio]{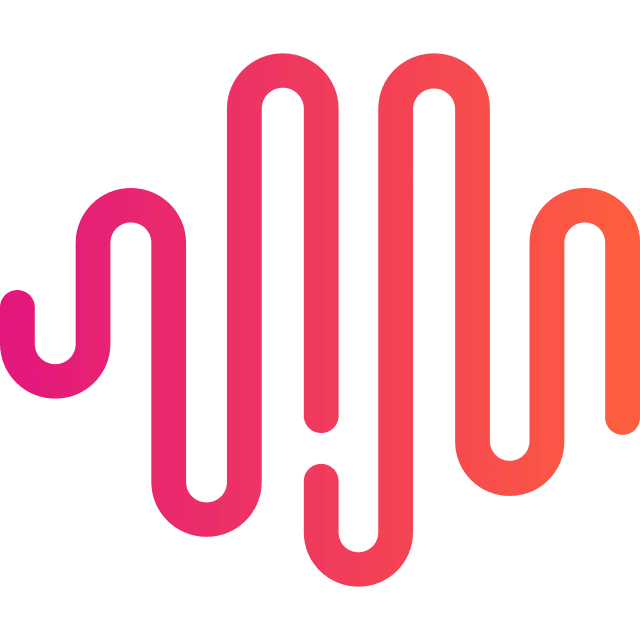}}\hspace{0.35em}MiniMax & \mbox{MiniMax-M3} & 5.61 & 34.50 & 26.18 & 45.59 & 36.63 & 11.70 & 35.57 & 27.97 \\
\raisebox{-0.15ex}{\includegraphics[width=1.1em,height=1.1em,keepaspectratio]{fig/provider_openai.png}}\hspace{0.35em}OpenAI & \mbox{GPT-6 Luna} & 20.83 & 54.69 & 30.06 & 56.54 & 51.09 & 17.50 & 49.82 & 40.08 \\
\raisebox{-0.15ex}{\includegraphics[width=1.1em,height=1.1em,keepaspectratio]{fig/provider_openai.png}}\hspace{0.35em}OpenAI & \mbox{GPT-5.5} & 20.55 & 66.23 & 30.29 & 60.34 & 53.84 & 17.50 & 44.88 & 41.95 \\
\raisebox{-0.15ex}{\includegraphics[width=1.1em,height=1.1em,keepaspectratio]{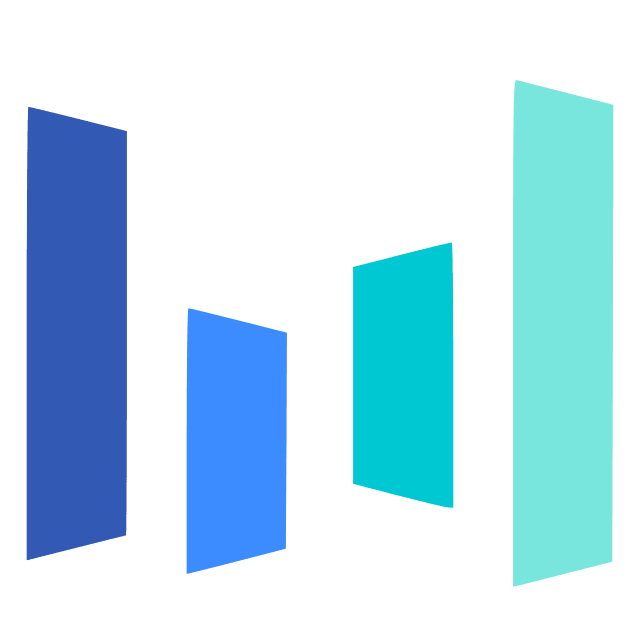}}\hspace{0.35em}ByteDance & \mbox{Doubao-Seed-2.1-Pro} & 18.20 & 64.58 & 39.88 & 62.87 & 59.79 & 29.75 & 64.00 & 48.44 \\
\raisebox{-0.15ex}{\includegraphics[width=1.1em,height=1.1em,keepaspectratio]{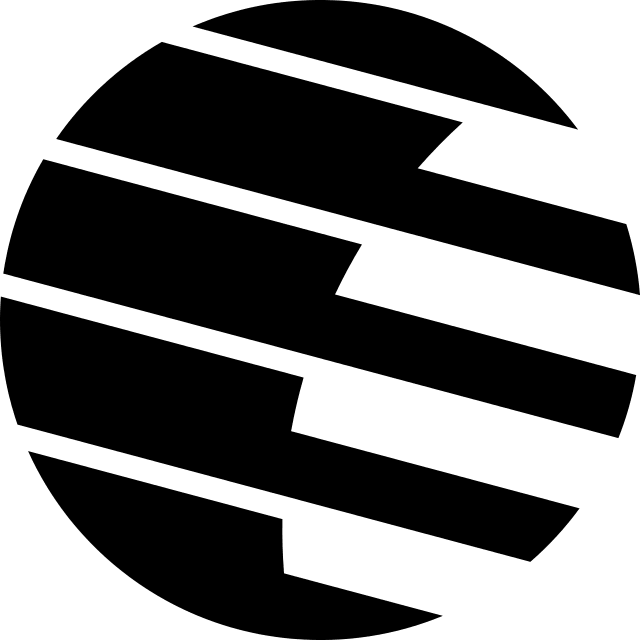}}\hspace{0.35em}Moonshot & \mbox{Kimi-K3} & 22.23 & 74.08 & 39.35 & 65.50 & 59.72 & 28.75 & 64.50 & 50.59 \\
\raisebox{-0.15ex}{\includegraphics[width=1.1em,height=1.1em,keepaspectratio]{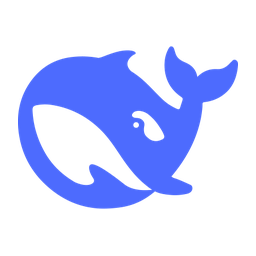}}\hspace{0.35em}DeepSeek & \mbox{DeepSeek-V4-Pro-0813} & 19.67 & 70.59 & 44.75 & 64.21 & 63.00 & 32.44 & 63.41 & 51.15 \\
\raisebox{-0.15ex}{\includegraphics[width=1.1em,height=1.1em,keepaspectratio]{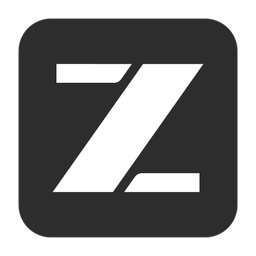}}\hspace{0.35em}Z.ai & \mbox{GLM-5.3-Flash} & 22.35 & 73.39 & 42.42 & 65.76 & 57.83 & 32.07 & 65.62 & 51.35 \\
\raisebox{-0.15ex}{\includegraphics[width=1.1em,height=1.1em,keepaspectratio]{fig/provider_openai.png}}\hspace{0.35em}OpenAI & \mbox{GPT-6.1 Sol} & 24.84 & 73.97 & 39.10 & 71.63 & 60.66 & 24.73 & 64.62 & 51.36 \\
\raisebox{-0.15ex}{\includegraphics[width=1.1em,height=1.1em,keepaspectratio]{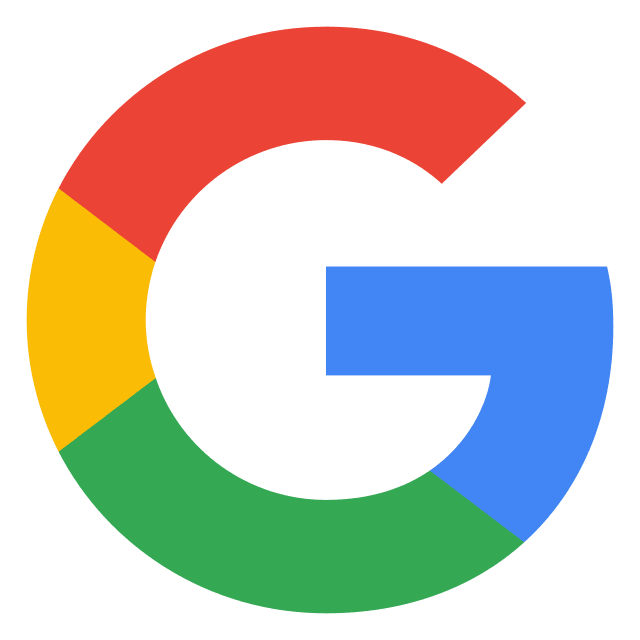}}\hspace{0.35em}Google & \mbox{Gemini-3.8-Flash} & \underline{25.70} & 72.47 & 40.51 & 65.49 & 60.56 & \underline{33.53} & 65.14 & 51.91 \\
\raisebox{-0.15ex}{\includegraphics[width=1.1em,height=1.1em,keepaspectratio,trim=35bp 56bp 35bp 9bp,clip]{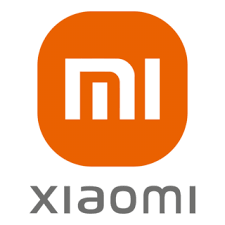}}\hspace{0.35em}Xiaomi & \mbox{MiMo-V2.6-Pro} & 24.60 & 73.94 & 44.73 & 65.86 & 61.79 & 27.93 & 67.25 & 52.30 \\
\raisebox{-0.15ex}{\includegraphics[width=1.1em,height=1.1em,keepaspectratio]{fig/provider_deepseek.png}}\hspace{0.35em}DeepSeek & \mbox{DeepSeek-V4.1-Flash} & 19.98 & 74.73 & 47.50 & 68.63 & 65.18 & 33.41 & 65.56 & 53.57 \\
\raisebox{-0.15ex}{\includegraphics[width=1.1em,height=1.1em,keepaspectratio]{fig/provider_alibaba.png}}\hspace{0.35em}Alibaba & \mbox{Qwen3.8-Max} & 24.09 & 73.44 & 46.98 & 67.56 & 66.25 & 31.10 & 68.64 & 54.01 \\
\raisebox{-0.15ex}{\includegraphics[width=1.1em,height=1.1em,keepaspectratio]{fig/provider_zai.png}}\hspace{0.35em}Z.ai & \mbox{GLM-5.3} & 23.78 & 75.02 & \underline{48.81} & 67.38 & 66.32 & 33.00 & 69.88 & 54.89 \\
\raisebox{-0.15ex}{\includegraphics[width=1.1em,height=1.1em,keepaspectratio]{fig/provider_openai.png}}\hspace{0.35em}OpenAI & \mbox{GPT-6 Astra} & \textbf{25.86} & \underline{77.50} & 42.11 & \underline{75.68} & 66.17 & 29.37 & \underline{74.12} & 55.83 \\
\raisebox{-0.15ex}{\includegraphics[width=1.1em,height=1.1em,keepaspectratio]{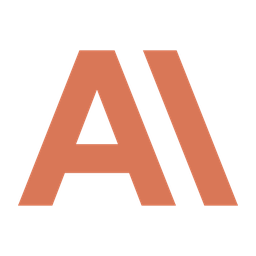}}\hspace{0.35em}Anthropic & \mbox{Claude Opus 5.5} & 21.94 & 74.98 & 46.75 & 71.70 & \textbf{73.22} & 33.27 & 72.89 & \underline{56.39} \\
\raisebox{-0.15ex}{\includegraphics[width=1.1em,height=1.1em,keepaspectratio]{fig/provider_google.png}}\hspace{0.35em}Google & \mbox{Gemini-4-Argon} & \textbf{25.86} & \textbf{81.12} & \textbf{50.37} & \textbf{76.83} & \underline{70.81} & \textbf{37.63} & \textbf{78.72} & \textbf{60.19} \\
\bottomrule
\end{tabular*}%
}
\end{table}

\subsection{Analysis}
\label{sec:analysis}

We examine how experience relates to task performance, how models adapt to environmental change, and how their tool use and experience maintenance respond.

\begin{figure*}[!t]
  \centering
  \includegraphics[width=\textwidth]{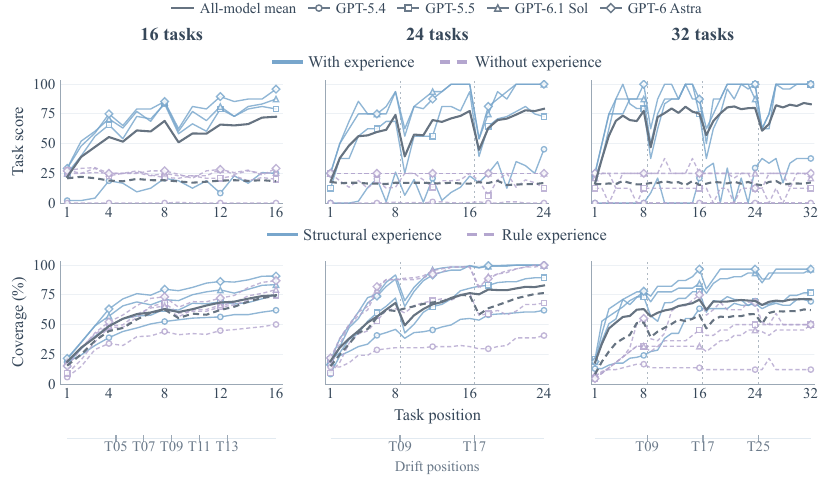}
  \caption{Task performance (top) and experience coverage (bottom) across task streams. Dark-gray curves average all 20 models, and markers distinguish four representative models. Bottom ticks show change positions, while vertical lines mark changes shared by all environments in a group.}
  \label{fig:rq1}
\end{figure*}

\subsubsection{How Do Experience and Task Performance Evolve Across Tasks?}
\label{sec:rq1}

To examine how experience relates to task performance, we track paired task scores and post-task structural and rule coverage across all 20 models and 18 environments. We group environments by stream length (16, 24, or 32 tasks), keeping the same environments equally weighted at every position.

\noindent\textbf{Gains accumulate across tasks.}
The upper row of Figure~\ref{fig:rq1} shows rising performance with experience across all three groups, while performance without experience remains low. In the 24- and 32-task groups, shared change onsets interrupt these gains, followed by recovery. The 16-task group combines different change positions, so its average has no single common disruption.

\noindent\textbf{Coverage losses differ by experience type.}
Coverage is scored against the current reference, so it can fall when previously correct content becomes outdated. In the 24-task group, performance drops at both T09 and T17. Structural coverage falls by 18.92 points at T09 while rule coverage stays comparatively stable. At T17, rule coverage falls by 15.97 points while structural coverage changes little. The two coverage measures distinguish mismatches that task scores alone cannot identify. Appendix~\ref{app:environment-trajectories} gives the individual environment curves.

\noindent\textbf{Coverage and performance can diverge.}
GPT-5.4 gains structural coverage across all three groups, yet its overall task performance remains far below the other displayed models. Coverage spans the full reference inventory, whereas successful delivery requires the relevant knowledge for a particular request and its effective use. Overall coverage therefore cannot substitute for task performance. Appendices~\ref{app:analysis-associations} and~\ref{app:analysis-requests} provide correlation and request-level analyses.

\begin{figure*}[!t]
  \centering
  \includegraphics[width=\textwidth]{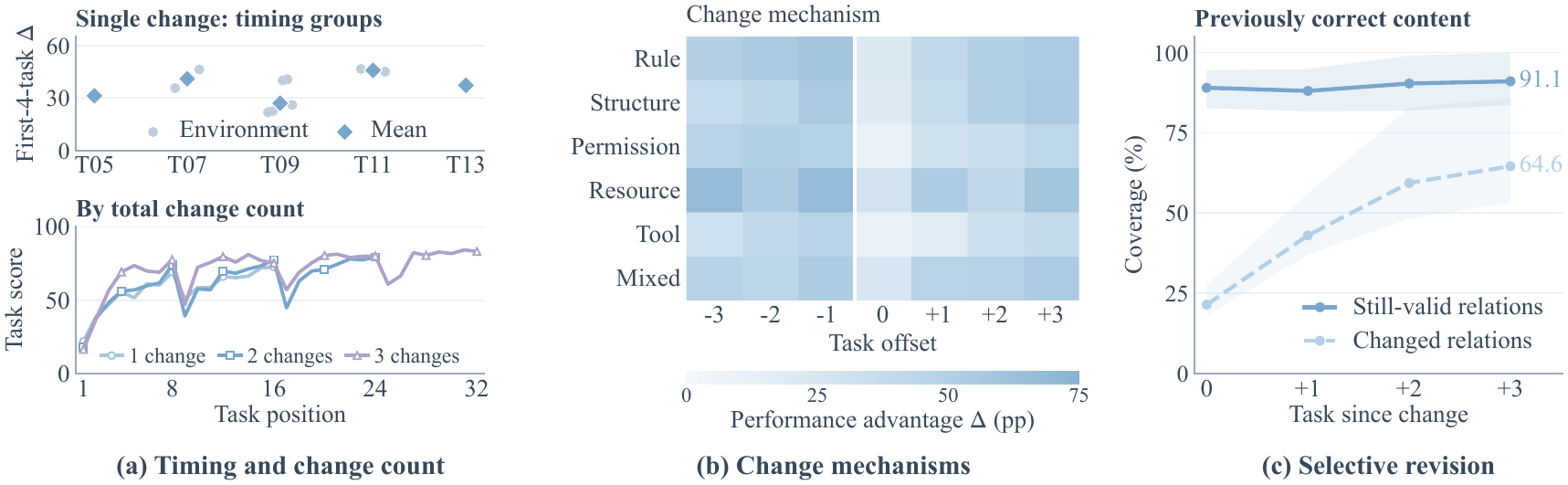}
  \caption{Adaptation to environmental change: (a) timing and change count, (b) performance advantage by change mechanism, and (c) revision and retention of previously correct experience.}
  \label{fig:rq2}
\end{figure*}

\subsubsection{How Do Models Adapt Across Different Change Contexts?}
\label{sec:rq2}

We examine performance recovery and selective experience revision under different change timings, counts, and mechanisms. Performance comparisons cover all 20 models, 18 environments, and 26 changes. Performance advantage denotes the task score with experience minus that without experience. In event-aligned comparisons, offset zero is the first affected task, and $+3$ is the fourth.

\noindent\textbf{Later events have higher average scores.}
Figure~\ref{fig:rq2}(a) shows no consistent increase in the first-four-task performance advantage when comparing single-change environments by onset position. The lower curves compare full task-score trajectories by change count. In two-change streams, mean with-experience scores over the first four affected tasks rise from 55.94 after the first event to 62.15 after the second. The corresponding means in three-change streams are 68.62, 70.50, and 72.53. Later events therefore coincide with higher average scores, although different tasks and change content prevent interpreting this as faster recovery.

\noindent\textbf{Similar drops, different recovery.}
In Figure~\ref{fig:rq2}(b), the resource group has the largest immediate loss of performance advantage, at 40.33 points. Structural and tool groups show similar losses of 35.98 and 36.63 points from the task immediately before change to the first affected task. By the fourth affected task, however, the structural group is only 0.16 points below its pre-change advantage, while the tool group remains 8.69 points below. Initial disruption thus does not capture the remaining loss after several tasks. These are descriptive comparisons across different environment groups.

\noindent\textbf{Revision remains incomplete.}
Figure~\ref{fig:rq2}(c) follows 487 cases across eight environments, each combining a model, reference experience unit, and change event to distinguish revision failure from missing prior knowledge. Each case correctly records all relations selected for updating and at least one still-valid relation before change. Over the first four affected tasks, replacement-content coverage rises from 21.47\% to 64.63\%, while still-valid coverage remains between 88.07\% and 91.07\%. Revision thus remains less complete than retention within this previously learned subset. The trajectories also reflect when tasks expose replacement evidence. Bands span the interquartile range of model means. Appendices~\ref{app:analysis-selective} and~\ref{app:revision-persistence} examine this exposure and subsequent persistence.

\subsubsection{How Do Agents Allocate Tool Use?}
\label{sec:rq3}

\begin{figure*}[!t]
  \centering
  \includegraphics[width=\textwidth]{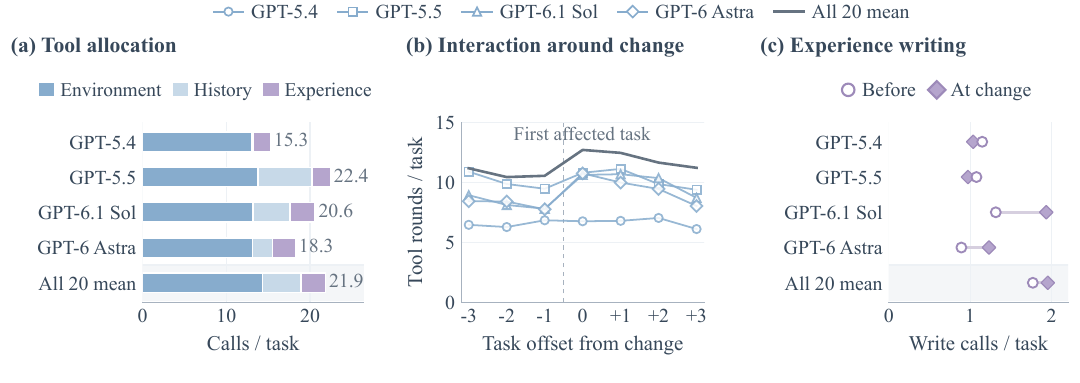}
  \caption{Tool use across 18 environments: (a) call allocation, (b) interaction rounds, and (c) experience write attempts around change. Representative models match Figure~\ref{fig:rq1}. The overall mean includes all 20 models.}
  \label{fig:rq3}
\end{figure*}

We analyze 7,040 with-experience tasks across all 20 models and 18 environments to examine interaction allocation and responses to change. History tools retrieve prior records, experience tools read or modify memory and skills, and environment tools query or act on the current state. We count explicit calls, including failures, and exclude background review. Changes are aligned at the first affected task. Appendix~\ref{app:tool-behavior} gives the counting and averaging procedures.

\noindent\textbf{Models differ in history retrieval.}
In Figure~\ref{fig:rq3}(a), the four displayed models make similar numbers of environment calls, ranging from 13.07 to 13.81 per task. History calls vary much more, from 0.14 for GPT-5.4 to 6.41 for GPT-5.5. Their differences in total activity therefore largely reflect consultation of prior records, rather than more requests to the current environment.

\noindent\textbf{Interaction peaks near change.}
A tool round comprises one assistant turn with one or more calls and their responses. In Figure~\ref{fig:rq3}(b), the overall mean rises from 10.72 rounds over the three preceding tasks to 12.70 at onset, then falls to 11.21 by the fourth affected task. All displayed models except GPT-5.4 show a similar increase near change, indicating that the aggregate response is not shared uniformly across models.

\noindent\textbf{Interaction and writing can diverge.}
Figure~\ref{fig:rq3}(c) compares write attempts over the three preceding tasks with the first affected task. Although GPT-5.5 uses more rounds at onset, its write attempts fall from 1.08 to 0.97 per task. GPT-6.1 Sol increases from 1.31 to 1.94, and GPT-6 Astra from 0.89 to 1.23. More interaction therefore need not coincide with more attempts to modify saved experience. These counts measure writing activity, not correct revision, which Figure~\ref{fig:rq2}(c) assesses separately.

\section{Related Work}
\label{sec:related}

\subsection{Experience-Based Agent Adaptation}
Experience-based adaptation methods differ in what they retain from interaction and how they update it. Reflexion and ExpeL turn feedback and trajectories into verbal reflections or reusable insights \citep{DBLP:conf/nips/ShinnCGNY23,DBLP:conf/aaai/Zhao0XLLH24}. Other methods retain procedures: Voyager stores executable skills \citep{DBLP:journals/tmlr/WangX0MXZFA24}, while Agent Workflow Memory abstracts recurring action sequences into reusable workflows \citep{DBLP:conf/icml/WangMFN25}. To keep guidance current, Dynamic Cheatsheet revises test-time memory as new evidence arrives \citep{DBLP:conf/eacl/SuzgunYBJZ26}. \resc{} complements these methods with an evaluation of how acquired experience is reused and maintained across tasks as local conditions change.

\subsection{Benchmarking Agent Memory and Experience}
Benchmarks study experience under supplied histories, interactive tasks, and changing conditions. LoCoMo, LongMemEval, and MemBench evaluate reasoning over supplied histories, with the latter two also assessing updates to remembered information \citep{DBLP:conf/acl/MaharanaLTBBF24,DBLP:conf/iclr/WuWYZCY25,DBLP:conf/acl/Tan000DD25}. MemoryAgentBench adds incremental learning and selective forgetting \citep{DBLP:journals/corr/abs-2507-05257}. Interactive benchmarks, including LifelongAgentBench and MemoryArena, evaluate experience use across tasks \citep{DBLP:journals/corr/abs-2505-11942,DBLP:journals/corr/abs-2602-16313}, while skill-oriented benchmarks study acquisition, refinement, and transfer \citep{DBLP:journals/corr/abs-2604-17308,DBLP:journals/corr/abs-2605-24117,DBLP:journals/corr/abs-2607-05202}. More recent benchmarks examine maintenance as information and environments change. AgingBench \citep{DBLP:journals/corr/abs-2605-26302} studies memory degradation and revision, including agent-managed workspace scenarios. EvoArena \citep{DBLP:journals/corr/abs-2606-13681} evaluates evolving terminal workflows, codebases, and preferences, and measures whether saved memory contains evidence of user preferences. In \resc{}, tasks reuse discoverable structural and rule experience, and local changes make selected parts outdated. We assess saved content alongside task outcomes to distinguish correction of affected relationships from retention of those that remain valid.

\section{Conclusion}

This paper introduced \resc{}, a benchmark for studying experience acquisition, reuse, and revision across 18 interactive environments. Its experience-first construction builds varied tasks around shared structural and rule experience, then introduces silent changes that invalidate selected content while leaving other content valid. Evaluations of 20 models show that retaining experience improves task performance, but rule experience is generally recorded less completely than structural experience. Further analyses show that agents preserve much of the still-valid content while leaving required updates incomplete, and can fail to complete requests even when the required experience is correctly recorded. More interaction after change does not always coincide with more explicit attempts to update saved experience. \resc{} offers a testbed for developing agents that selectively revise outdated experience and effectively apply what they retain.

% Submission statements are excluded from the main-text page count.
\label{main:end}

\section*{Acknowledgments}
This research was financially supported by the Open Research Fund from Guangdong Laboratory of Artificial Intelligence and Digital Economy (SZ), under Grant No. GML-KF-26-06.

\clearpage
% \section*{Reproducibility statement}
% Section~\ref{sec:construction} describes benchmark construction and verification. Section~\ref{sec:metrics} specifies the evaluation protocol, metric definitions, and aggregation rules. Section~\ref{sec:analysis} details the analyses. Appendix~\ref{app:environment-inventory} lists the environments and change schedules. Appendix~\ref{app:evaluation-details} specifies agent inputs, evaluation settings, and metric implementation. Appendices~\ref{app:analysis-scope} and \ref{app:experience-cases} specify the quantitative analysis and source-grounded semantic audit.

% \section*{AI use statement}
% We used generative AI as a supporting tool for benchmark development, analysis, and manuscript preparation. The authors take full responsibility for the study's claims, factual accuracy, interpretation of results, and final text.

% \section*{Ethics statement}
% \resc{} uses bounded simulated environments. Its results do not establish the safety or reliability of autonomous deployment in real-world or safety-critical settings, which requires separate validation.

\bibliography{iclr2027_conference}
\bibliographystyle{iclr2027_conference}

\clearpage
\appendix

\section*{Appendix Contents}
\begingroup
\small
\newcommand{\appentry}[2]{\noindent\hyperref[#1]{\ref*{#1}\quad #2}\hfill\pageref{#1}\par\vspace{3pt}}
\appentry{app:environment-inventory}{\textbf{Benchmark Environments and Changes}}
\appentry{app:evaluation-details}{\textbf{Evaluation Details}}
\hspace*{1em}Models, paired streams, experience access, execution, and metrics\par\vspace{3pt}
\appentry{app:crossjudge}{\hspace{.6em}Cross-judge consistency}
\appentry{app:analysis}{\textbf{Supplementary Analyses}}
\appentry{app:experience-performance}{\hspace{.6em}Experience and task performance}
\hspace*{2em}Environment trajectories, request outcomes, associations, and acquisition\par\vspace{3pt}
\appentry{app:change-maintenance}{\hspace{.6em}Environmental change and experience maintenance}
\hspace*{2em}Selective revision and persistence\par\vspace{3pt}
\appentry{app:analysis-actions}{\hspace{.6em}Interaction and action allocation}
\appentry{app:tool-behavior}{\hspace{1.2em}Tool use across environments}
\appentry{app:request-action-paths}{\hspace{1.2em}Request-level action paths}
\appentry{app:experience-cases}{\textbf{Illustrative Case Studies}}
\hspace*{1em}Local revision and reuse, and use of saved experience under a budget\par
\endgroup
\clearpage

% Requires array, booktabs, and longtable.
% Evidence: the environment catalog in main_before_appendix_reorganization.tex.
\section{Benchmark Environments and Changes}
\label{app:environment-inventory}
\label{app:environment-catalog}

This appendix expands the environment inventory in Section~\ref{sec:statistics}. Table~\ref{tab:environment-catalog} lists task goals, stream lengths, and tool counts for all 18 environments. Counts exclude the eight shared experience and history tools. Table~\ref{tab:environment-changes} lists all 26 scheduled changes in the evaluated versions. Each change position denotes the first affected task.

\subsection{Task Goals and Tools}

\begin{table*}[!t]
  \centering
  \fontsize{9}{10.5}\selectfont
  \setlength{\tabcolsep}{4pt}
  \renewcommand{\arraystretch}{1.12}
  \caption{Task goals, stream lengths ($T_e$), and tool counts across 18 environments.}
  \label{tab:environment-catalog}
  \label{tab:environment-inventory}
  \begin{tabular}{@{}>{\raggedright\arraybackslash}p{0.18\textwidth}>{\centering\arraybackslash}p{0.04\textwidth}>{\centering\arraybackslash}p{0.07\textwidth}>{\raggedright\arraybackslash}p{\dimexpr0.71\textwidth-6\tabcolsep\relax}@{}}
    \toprule
    Environment & $T_e$ & Tools & Task goal \\
    \midrule
    API Reliability & 16 & 5 & Restore failed provider requests to durable accepted results without duplicate effects. \\
    Backup and Recovery & 16 & 14 & Create active asset-protection assignments with daily schedules, activated policies, and site-compatible repositories. \\
    Carbon Credits & 32 & 7 & Complete carbon-credit retirements using currently eligible holdings and required audit evidence. \\
    Cinema Booking & 24 & 9 & Complete paid bookings satisfying the requested film, venue, showtime, ticket type, and seat zone, using eligible pricing and available seats. \\
    Cloud Data Warehouse & 24 & 5 & Commit governed query deliveries with materialized tables, completed exports, valid lane authorities, and applicable approved controls. \\
    Course Registration & 16 & 15 & Create active student registrations in the requested sections and processing lanes under the applicable policy. \\
    \addlinespace
    Emergency Response & 16 & 8 & Dispatch authorized, available resources. Release each assignment and verify that the resource is available again. \\
    Event Ticketing & 24 & 15 & Complete paid orders with the requested admission class, two distinct available seats, the eligible house offer, and issued tickets. \\
    Filesystem & 16 & 7 & Materialize manifest indexes in their exact eligible delivery lanes under current workspace and delivery-policy bindings. \\
    Kubernetes Operations & 24 & 7 & Publish deployments and services with replicas in the required zones, current namespace-assigned worker pools, and compliant rollout/service settings. \\
    Medical Appointments & 16 & 4 & Schedule policy-compliant follow-ups linked to completed encounters, using currently eligible slots. \\
    MES Order Closeout & 16 & 10 & Close production orders after satisfying material-issue, operation, equipment, and quality preconditions. \\
    \addlinespace
    Oracle Clinical One & 16 & 3 & Correct completed forms while preserving the supplied values and policy-compliant audit records. \\
    Railway Ticketing & 32 & 18 & Issue new confirmed bookings matching the passenger, route, date, seat class, and requested seat position. \\
    REDCap & 16 & 13 & Resolve record work packets by verifying corrected fields, closing queries, completing both forms, and locking the record event. \\
    Robot Logistics & 16 & 14 & Deliver parcels for priority-ordered work orders, close each order, and return the empty robot to the service hub. \\
    SAP S/4HANA & 16 & 17 & Authorize purchase-to-pay routes, receive outstanding quantities, create receipt-backed supplier invoices, and post them in open company periods. \\
    Supply Chain Finance & 16 & 11 & Finance eligible invoice balances under current program authority, then complete settlement and matched reconciliation. \\
    \bottomrule
  \end{tabular}
\end{table*}

\clearpage
\subsection{Complete Change Catalog}

The catalog uses the six change mechanisms defined in Section~\ref{sec:statistics}. We classify events by their effect on environment operations, rather than the names of modified fields. For example, Cinema's replacement pricing plans revise rules, whereas Railway's replacement train services change available resources.

\begingroup
\fontsize{9}{10.5}\selectfont
\setlength{\tabcolsep}{4pt}
\renewcommand{\arraystretch}{1.12}
\setlength{\LTcapwidth}{\textwidth}
\setlength{\LTleft}{0pt}
\setlength{\LTright}{0pt}
\begin{longtable}{@{}>{\raggedright\arraybackslash}p{0.18\textwidth}>{\centering\arraybackslash}p{0.055\textwidth}>{\raggedright\arraybackslash}p{0.11\textwidth}>{\raggedright\arraybackslash}p{\dimexpr0.655\textwidth-6\tabcolsep\relax}@{}}
  \caption{All 26 scheduled changes. Task denotes the first affected task position.}
  \label{tab:environment-changes}
  \label{tab:environment-changes-a}
  \label{tab:environment-changes-b}
  \label{tab:environment-changes-c}\\
  \toprule
  Environment & Task & Mechanism & Implemented change \\
  \midrule
  \endfirsthead
  \multicolumn{4}{l}{\textit{Table~\thetable\ (continued)}}\\
  \toprule
  Environment & Task & Mechanism & Implemented change \\
  \midrule
  \endhead
  \midrule
  \multicolumn{4}{r}{\textit{Continued on the next page}}\\
  \endfoot
  \bottomrule
  \endlastfoot
  API Reliability & 9 & Rule & Revise recovery profiles for four operation-specific cases: queued replays, provider errors, timeouts, and rate limits. Retain the retry action. \\
  \addlinespace[2pt]
  Backup and Recovery & 5 & Rule & Replace protection-route and repository-profile requirements for critical/restricted, standard/confidential, archive/internal, and regulated/public service-tier/data-class pairs. \\
  \addlinespace[2pt]
  Carbon Credits & 9 & Rule & Revise project eligibility and claim protocols for Gold Standard customer allocations and national voluntary claims. Retain holdings. \\
  Carbon Credits & 17 & Rule & Revise an allocation policy and again amend the national voluntary-claim policy. Retain the earlier Gold Standard allocation rule. \\
  Carbon Credits & 25 & Rule & Revise Gold Standard allocation and national voluntary-claim policies with new protocols. Retain the allocation rule changed at task 17. \\
  \addlinespace[2pt]
  Cinema Booking & 9 & Resource & Cancel showtimes in three screening groups and open replacements. Transfer screening-specific pricing assignments without changing fare terms. \\
  Cinema Booking & 17 & Rule & Change accepted checkout channels in three pricing-policy cells. Retire old pricing plans and bind screenings to successor plans. \\
  \addlinespace[2pt]
  Cloud Data Warehouse & 9 & Structure & Rebind delivery lanes to new routing/lineage authorities and destination tuples. Preserve delivery policies and tools. \\
  Cloud Data Warehouse & 17 & Rule & Introduce region-qualified sensitive-data controls: sensitive EU destinations require compressed-Parquet snapshots. Retain domestic profiles. \\
  \addlinespace[2pt]
  Course Registration & 9 & Tool-path & Change selected workflows to queue then promote, provisionally register then finalize grading basis, or review the profile then register directly. \\
  \addlinespace[2pt]
  Emergency Response & 9 & Permission & Replace recognized responsible/requesting agencies in selected authority domains. Previous requesters no longer authorize dispatch. \\
  \addlinespace[2pt]
  Event Ticketing & 9 & Structure & Rebind ticket types to price tiers and migrate event, performance, and ticket-type identifiers. Preserve seats, purchases, and commercial entitlements. \\
  Event Ticketing & 17 & Rule & Revise ticket-type eligibility for commercial discount/offer profiles. Retain the migrated catalog identifiers. \\
  \addlinespace[2pt]
  Filesystem & 7 & Structure & Replace selected destination-lane nodes with successor handles for the same delivery purposes. In Grove, only the amber lane changes. \\
  \addlinespace[2pt]
  Kubernetes Operations & 9 & Structure & Reassign namespace-zone allocations to other existing worker pools and rekey namespaces/references. Preserve workloads, capacity, and policies. \\
  Kubernetes Operations & 17 & Rule & Revise rollout strategy, service type, external-DNS, and NodePort requirements in two policy cells. Retain namespace-to-pool assignments and tools. \\
  \addlinespace[2pt]
  Medical Appointments & 11 & Rule & Revise care-plan-update routing for standard, chronic-care, senior-care, and maternity patients, changing continuity/team and in-person/virtual requirements. \\
  \addlinespace[2pt]
  MES Order Closeout & 7 & Mixed & Quarantine old material lots and allocate replacements. Separately revise material-issue quantities for four closeout templates. \\
  \addlinespace[2pt]
  Oracle Clinical One & 11 & Rule & Revise correction audit-mode/reason profiles for screening/transcription, scheduled/clarification, completion/unit normalization, and dispensation/coding alignment visit/correction pairs. \\
  \addlinespace[2pt]
  Railway Ticketing & 9 & Resource & Cancel generation-0 services for the first two route groups (A). Activate generation-1 replacements with their own seat inventories. \\
  Railway Ticketing & 17 & Resource & Replace generation-0 services for the other two route groups (B) with generation-1 services and their own seat inventories. Retain group A services. \\
  Railway Ticketing & 25 & Resource & Replace group A's generation-1 services with generation-2 services and their own seat inventories. Retain group B services and issued tickets. \\
  \addlinespace[2pt]
  REDCap & 9 & Permission & Deactivate old project-user authorities and activate successors. Transfer query assignments and execution authority to them. \\
  \addlinespace[2pt]
  Robot Logistics & 9 & Mixed & Rotate local intake/transfer policies and relocate selected groups of parcels,  updating parcel locations, future source rooms, room settings, and navigation/pick locations. \\
  \addlinespace[2pt]
  SAP S/4HANA & 9 & Tool-path & Switch the authorization tool and route reference for four short-payment-horizon purchasing routes. Retain receipt, invoicing, and posting workflows. \\
  \addlinespace[2pt]
  Supply Chain Finance & 13 & Permission & Replace the financing authority bundle: funding scope, approval chain, delegation, authority namespace, and submission channel. Retain the financing tool. \\
\end{longtable}
\endgroup

\clearpage
\section{Evaluation Details}
\label{app:evaluation-details}
\label{app:agent_experience_setup}
\label{app:agent-inputs}

\subsection{Experimental Setup}
\label{app:evaluated-models}
\label{app:model-configurations}
\label{app:paired-streams}
\label{app:execution-details}
We evaluate the 20 models in Table~\ref{tab:main-results} on the same 18 versioned environments. Each model completes 352 tasks per condition, covering 26 environmental changes. We use the maximum supported reasoning effort where configurable. Explicit sampling settings use temperature 1 and top-$p$ 0.95; unsupported or omitted fields follow provider defaults. Foreground execution allows 100 agent iterations per task and up to 65,536 output tokens per response. An iteration may contain multiple tool calls. Environment-specific resource budgets apply separately.

The Hermes memory and skill interfaces are pinned to revision \texttt{345cd2b057a45}. Memory and skills are limited to 16,000 and 100,000 characters, respectively. Agents receive a memory snapshot and skill index at task start and can retrieve full skills on demand. Four read-only history tools retrieve completed-task records through lexical matching and pagination, with a shared limit of 30 calls per task. Structural and rule experience are classified by content, regardless of where they are stored. No reference experience is preloaded.

Both conditions start from the same environment state and use a fresh conversation for each task. With-experience streams retain memory, skills, and retrievable history. Without-experience streams clear these assets while retaining their own environment state. The two states then evolve independently through agent actions. Each model--environment pair has one stream per condition. GPT-5.5 also supported environment development. Agents can update experience during execution. Hermes skill review can run after ten tool iterations and a completed, uninterrupted response, with a separate limit of 100 iterations and 600,000 cumulative input tokens. It uses the same model, cannot access environment tools, and finishes before grading, so the current task's score cannot supervise review. Later history retrieval may include permitted outcome feedback, but never private reference answers.

\subsection{Scoring and Aggregation}
\label{app:metric-definitions}
\label{app:recovery-eligibility}
BP and EP average all task positions, including the initial task. DP uses the first affected task after each change and includes adaptation within that task. SC and RC average binary claim scores within reference units, units within each experience type, and checkpoints within each environment. References follow the current environment version and include content not yet encountered by the agent.

GPT-6 Astra evaluates one complete reference unit against the complete saved memory and skill corpus per request, using medium reasoning effort and a 14,000-token output limit. Model-identifying metadata is withheld. Each claim receives credit only when all applicable rubric checks pass. Equivalent wording and supported deductions are accepted, while unresolved contradictions receive no credit. We validate the structured responses and cited evidence before computing scores. These metrics measure reference coverage rather than the precision of all saved content.

For each affected unit, UC measures coverage of the current update targets after the first affected task. RT measures the first checkpoint at which all update targets and the unit's previously correct, still-valid claims are correct together. Update targets are shared across models, while the retained set depends on pre-change credit. Both target sets remain fixed within an $L$-task window starting at the first affected task and ending before the next change or at stream completion. Only units with nonempty update targets participate. Completion at its first or last task scores 100 or $100/L$, respectively, and no completion scores zero. An empty retained set reduces the criterion to update completion. RT records first completion, not its subsequent persistence, and reflects the opportunities provided by the task schedule.

All metrics weight environments equally. DP averages events within environments. UC and RT average affected units within events, then events within environments. Avg.\ averages the seven unrounded metrics. Longer streams and larger inventories do not receive more weight.

\label{app:analysis-support}
Supplementary analyses use $W_{met}$ and $B_{met}$ for task scores with and without experience, and $S_{met}$ and $R_{met}$ for post-task structural and rule coverage. The difference $\Delta_{met}=W_{met}-B_{met}$ is distinct from DP. Position curves keep environment membership fixed within each stream-length group and equally average environments and models. Model interquartile ranges describe variation across models, not uncertainty across repeated runs. Inapplicable observations are excluded rather than scored as zero.

\subsection{Cross-Judge Consistency}
\label{app:crossjudge}
We rejudge all 7,040 checkpoints from 20 models and 18 environments with GPT-6.1 Sol, holding saved content, references, rubrics, prompts, and inference settings fixed. The second judge does not see the first judge's decisions. The comparison covers 121,717 distinct unit-level requests and 433,850 claim decisions. All requests specify Sol without fallback. Of valid responses, 81.0\% carry an explicit Sol deployment label and 19.0\% an opaque provider identifier.

\begin{figure}[!t]
\centering
\includegraphics[width=\linewidth]{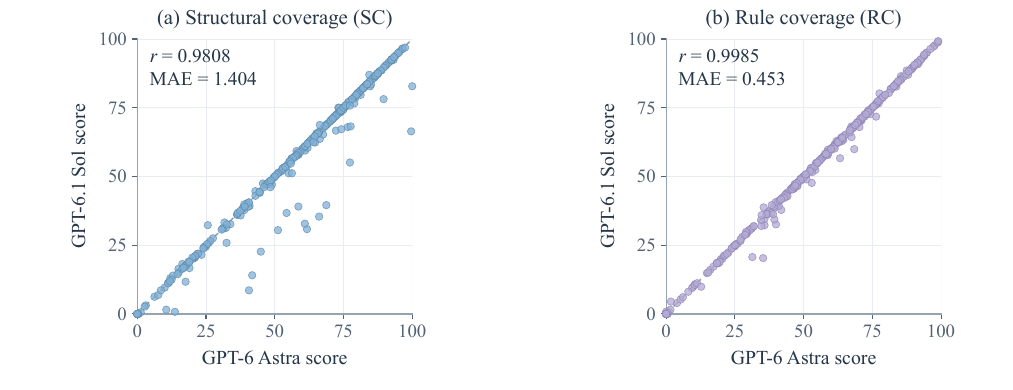}
\caption{Agreement between GPT-6 Astra and GPT-6.1 Sol experience scores across 360 model--environment pairs per panel. The dashed line denotes equal scores.}
\label{fig:crossjudge}
\end{figure}

The judges agree on 98.79\% of structural and 99.24\% of rule claim decisions. Across 360 model--environment pairs, Pearson correlations for SC, RC, UC, and RT range from 0.9808 to 0.9985, as shown in Figure~\ref{fig:crossjudge}. The aggregate structural--rule coverage gap remains positive with both judges (4.79 points for Astra and 3.86 for Sol). Agreement is weaker in some environments, particularly SAP structural coverage. These results support consistency under the shared rubric, rather than independent validation of judgment accuracy.

% Detailed cross-judge table retained for reference.
% \begin{table}[!t]
% \centering
% \caption{Cross-judge agreement under identical rubrics. $r$ is Pearson correlation, $\rho$ is Spearman rank correlation, and MAE is mean absolute error in score points.}
% \label{tab:crossjudge}
% \small
% \setlength{\tabcolsep}{6pt}
% \renewcommand{\arraystretch}{1.1}
% \begin{tabular*}{\linewidth}{@{\extracolsep{\fill}}lrrrrr@{}}
% \toprule
% & \multicolumn{2}{c}{360 model--environment pairs} & \multicolumn{3}{c}{20 model averages} \\
% \cmidrule(lr){2-3}\cmidrule(lr){4-6}
% Metric & $r$ & MAE & $r$ & $\rho$ & MAE \\
% \midrule
% SC & 0.9808 & 1.404 & 0.9984 & 0.9805 & 1.231 \\
% RC & 0.9985 & 0.453 & 0.9998 & 0.9940 & 0.337 \\
% UC & 0.9905 & 0.469 & 0.9963 & 0.9759 & 0.469 \\
% RT & 0.9973 & 0.476 & 0.9995 & 0.9865 & 0.423 \\
% \bottomrule
% \end{tabular*}
% \end{table}

\section{Supplementary Analyses}
\label{app:analysis}
\label{app:analysis-scope}
\label{app:quantitative-analysis}
This section provides additional evidence for the three analyses in Section~\ref{sec:analysis}: experience and task performance, adaptation to environmental change, and action allocation. Complete-stream analyses cover all 20 models and 18 environments. Request-level analyses use Filesystem and Oracle, where verified mappings connect required experience, saved content, and observable actions.

\subsection{Experience and Task Performance}
\label{app:experience-performance}
\label{app:retained-lifecycle-evidence}
\subsubsection{Complete Per-Environment Trajectories}
\label{app:environment-trajectories}
\begin{figure}[p]
  \centering
  \includegraphics[width=\textwidth]{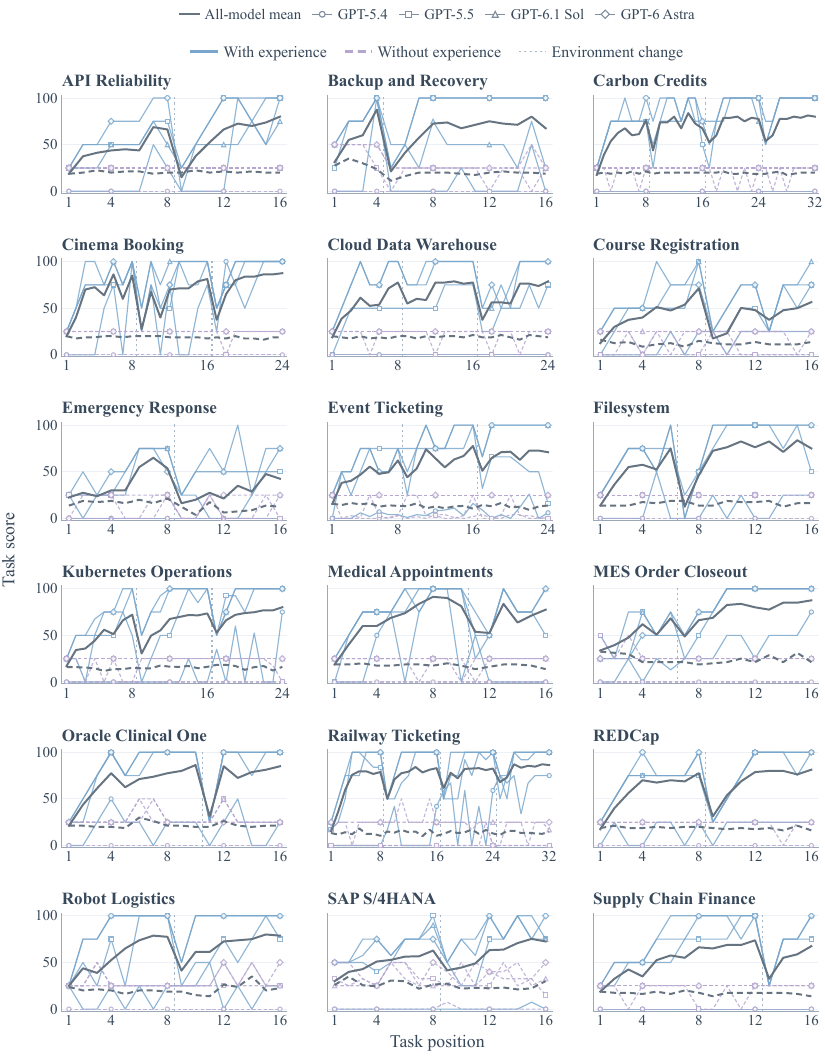}
  \caption{Task performance in all 18 environments. Solid and dashed curves show performance with and without experience. Vertical lines mark changes.}
  \label{fig:app-environment-performance}
\end{figure}

\begin{figure}[p]
  \centering
  \includegraphics[width=\textwidth]{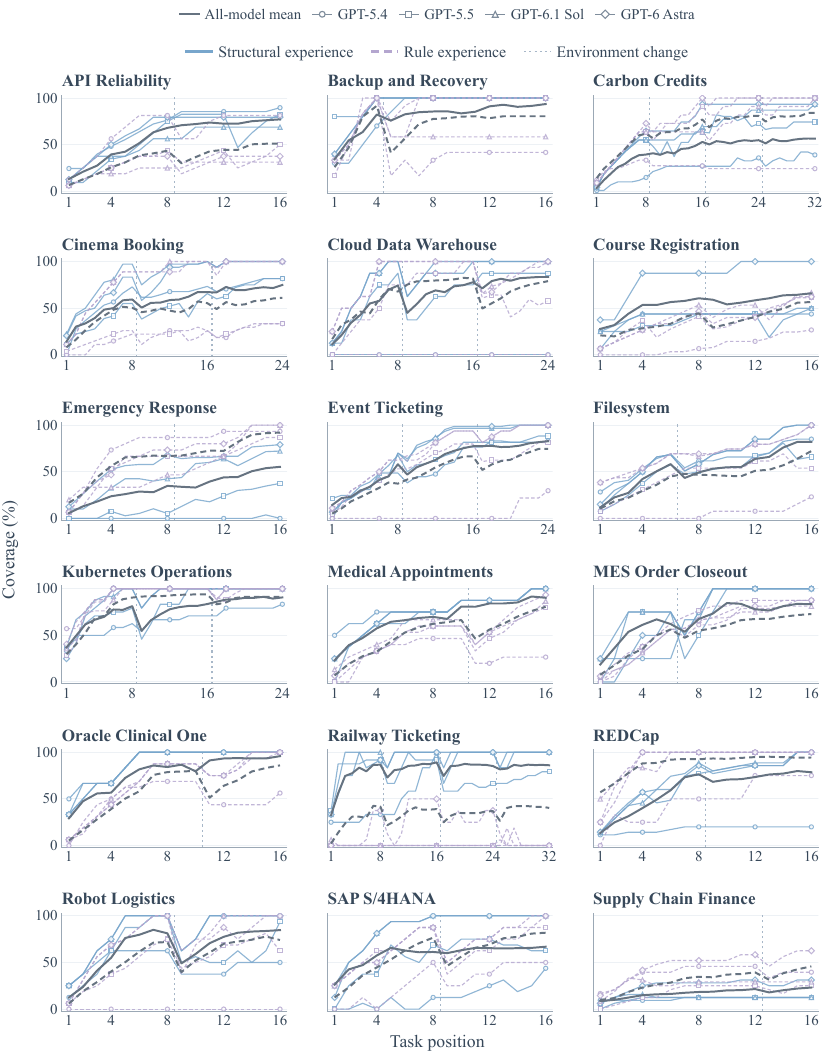}
  \caption{Post-task experience coverage in all 18 environments. Solid and dashed curves show structural and rule coverage. Vertical lines mark changes.}
  \label{fig:app-environment-experience}
\end{figure}

Figures~\ref{fig:app-environment-performance} and~\ref{fig:app-environment-experience} show the environment-level variation behind the aggregate trends. They plot complete task streams, with markers for GPT-5.4, GPT-5.5, GPT-6.1 Sol, and GPT-6 Astra as in Figure~\ref{fig:rq1}. Dark-gray curves equally average the same 20 models at every position. The first figure shows task performance in both conditions. The second shows post-task structural and rule coverage in the with-experience condition, measured against the current reference. Vertical lines immediately precede each change's first affected task.

\subsubsection{Task-Required Content and Request Outcomes}
\label{app:analysis-requests}
To relate saved experience to delivery, we align each request's required structural and rule bindings with saved claims, submitted arguments, and delivered artifacts. Figure~\ref{fig:app-request-actions}(a) summarizes the first-submission paths. The trace cohort covers 640 with-experience tasks, 2,560 requests, and 8,826 foreground tool calls with matching responses across all 20 models in Filesystem and Oracle. We select these environments for their verified mappings, rather than as a random sample.

A request at task $t$ has \emph{complete prior experience} when every required structural and rule claim is credited at checkpoint $t-1$ under the same reference version. Credit requires all required and applicable rubric checks to pass. Prior experience is incomplete if any mapped claim lacks credit. The 320 requests on initial and change-onset tasks have no same-version predecessor, so their prior status is unknown. The remaining 2,240 requests comprise 1,014 complete-prior and 1,226 incomplete-prior requests.

We compare complete and incomplete priors within each model's tasks that contain both groups and have resolved submission attribution. This yields 326 model/environment/task cells, 1,304 requests (625 complete-prior and 679 incomplete-prior), and 36 model/environment cells across 18 models, without selecting on delivery. We compute each group's rate within a task, then average over eligible tasks, models within each environment, and the two environments equally. Both groups share task cells and outer weights. Eligible tasks vary across models: 5--14 in Filesystem and 5--10 in Oracle.

\begin{figure}[!t]
  \centering
  \includegraphics[width=.86\textwidth]{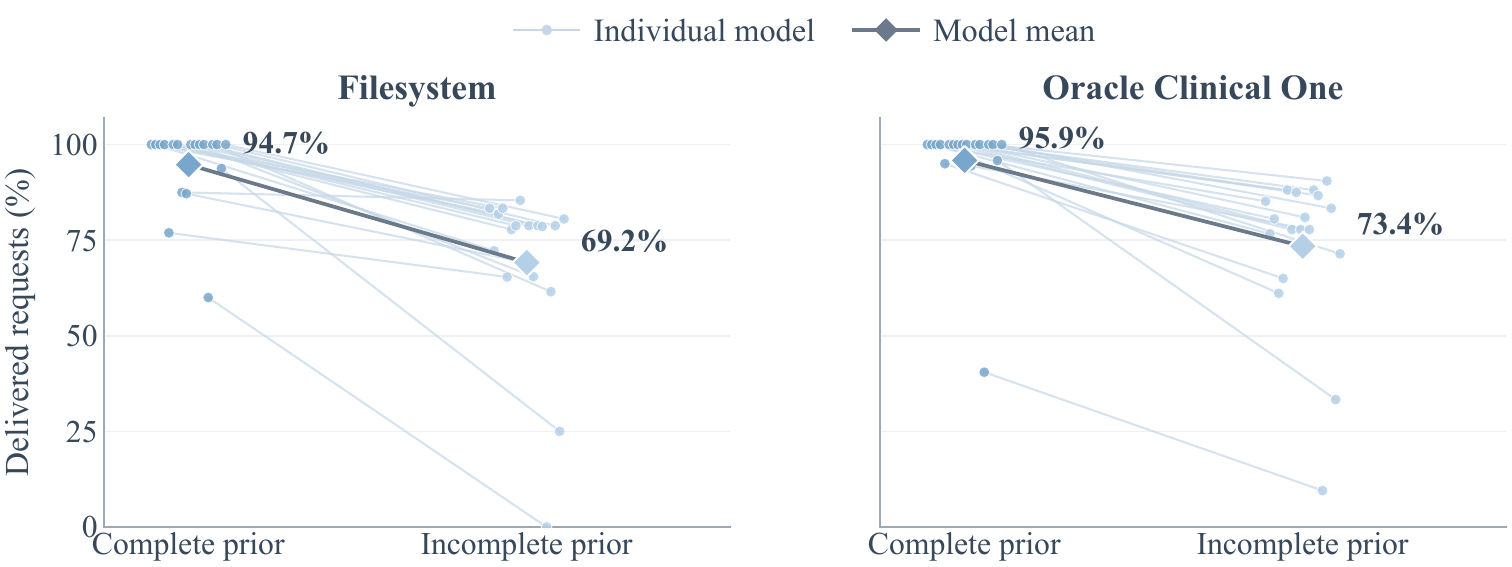}
  \caption{Delivery by prior-content completeness within matched tasks. Lines connect each model's two rates, and diamonds mark equal-model means.}
  \label{fig:app-task-matched-delivery}
\end{figure}

Figure~\ref{fig:app-task-matched-delivery} shows higher delivery with complete prior experience in all 36 model/environment cells. Complete- and incomplete-prior delivery rates average 94.74/69.16\% in Filesystem and 95.88/73.40\% in Oracle. The cross-environment rates are 95.31\% and 71.28\%, a 24.03-point gap. Requests without a linked submission account for 19.42 points, submissions that never become compatible for 4.58, and compatible submissions without delivery for 0.03. The gap therefore mainly concerns whether requests reach submission.

Table~\ref{tab:analysis-stages} separates four mutually exclusive outcomes. Compatibility requires all frozen structural and rule bindings, while delivery is assessed from the resulting task artifact. An incompatible submission can be corrected before delivery. This decomposition is descriptive: matched requests can still differ in targets, ordering, remaining budget, and within-task discovery. In raw counts, 21 complete-prior requests fail delivery and 458 incomplete-prior requests succeed. Complete saved prior experience is thus neither necessary nor sufficient for delivery in this cohort.

\begin{table*}[!t]
  \centering
  \caption{Task-matched request outcome rates (\%) by completeness of prior experience.}
  \label{tab:analysis-stages}
  \fontsize{9}{11}\selectfont
  \begin{tabular}{lrr}
    \toprule
    Outcome & Complete prior & Incomplete prior \\
    \midrule
    No linked submission & 3.67 & 23.09 \\
    Submitted, never compatible & 0.63 & 5.21 \\
    Compatible, not delivered & 0.40 & 0.42 \\
    Delivered & 95.31 & 71.28 \\
    \bottomrule
  \end{tabular}
\end{table*}

\subsubsection{Lagged Experience--Performance Associations}
\label{app:analysis-associations}
We next examine whether prior coverage remains associated with performance after accounting for model and task variation. We pair checkpoint $t-1$ with task $t$ only within the same reference version. Excluding 360 initial and 520 change-onset observations leaves 6,160 observations at 308 environment/task positions. Unlike the request-specific prerequisites above, these predictors cover the full reference inventory.

\begin{figure*}[!t]
  \centering
  \includegraphics[width=\textwidth]{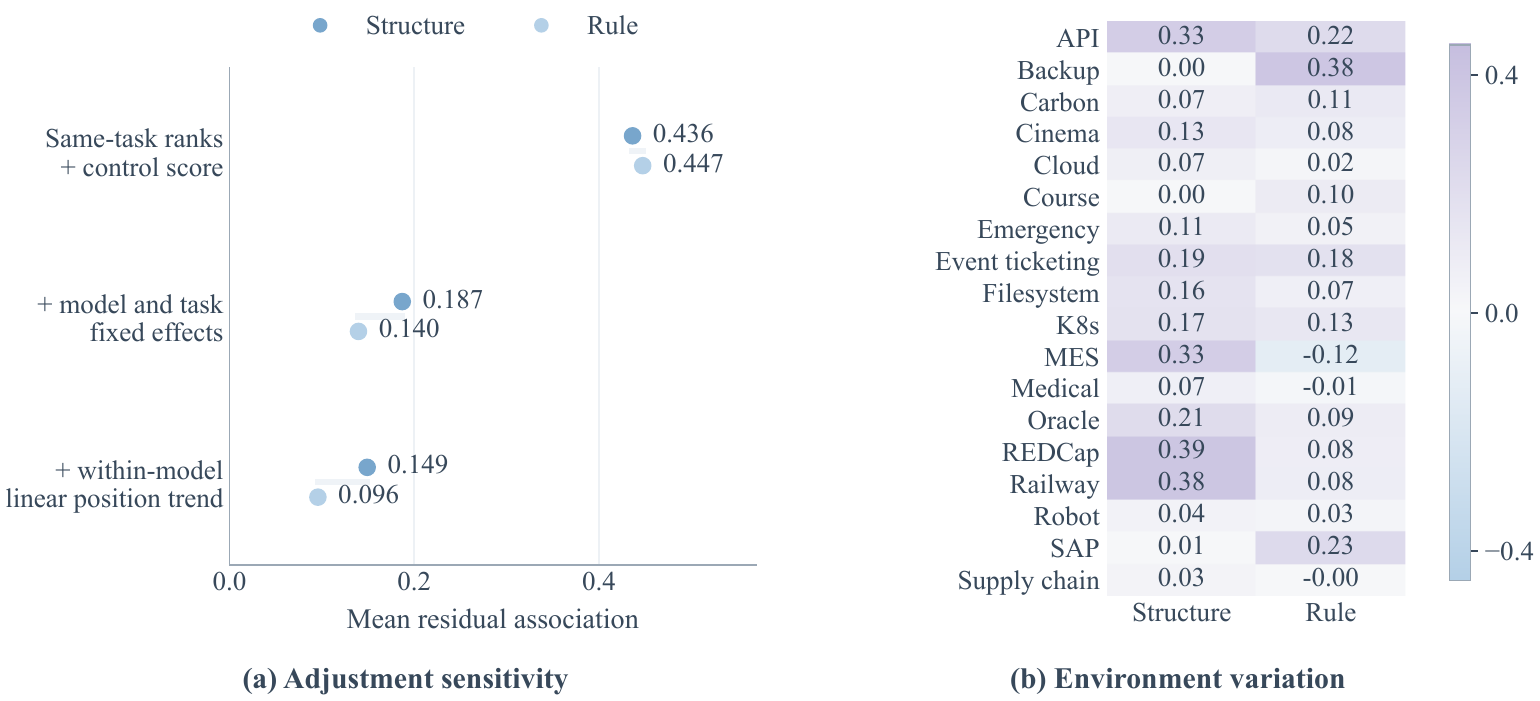}
  \caption{Prior-coverage correlations with current performance over 6,160 same-version observations. (a) Equal-environment means under successive adjustments. (b) Fully adjusted per-environment coefficients.}
  \label{fig:app-associations}
\end{figure*}

Across model means, structural and rule coverage correlate strongly with with-experience performance (Pearson $r=0.956/0.975$), but also with baseline performance ($0.898/0.928$). The coverage measures correlate at $0.965$. We rank models within each task, average tied ranks, and adjust coverage and with-experience performance for contemporaneous baseline performance within each environment. We compute partial rank correlations per environment and average them equally. Figure~\ref{fig:app-associations} then adds model/environment and environment/task fixed effects, followed by model/environment-specific linear position trends, using the same observations.

The structural/rule associations fall from $0.436/0.447$ after baseline adjustment to $0.187/0.140$ with fixed effects and $0.149/0.096$ with position trends. The fully adjusted raw-score counterparts are $0.137/0.124$. Much of the aggregate association thus overlaps with differences among models, tasks, and stream progression. These descriptive adjustments do not identify a causal effect of saved experience. Residualizing either $W$ or $W-B$ on $B$ yields the same residual, so the two adjusted results provide the same evidence.

\subsubsection{Acquisition and Loss of Previously Credited Content}
\label{app:formation-decomposition}
We distinguish claims never credited from claims credited earlier but not currently, using all 20 models in the 15 environments unchanged through T08. The cohort and reference version remain fixed. At T04, the structural--rule coverage gap is 4.06 points: never-credited content contributes 4.725 points, while lost credit contributes $-0.662$. By T08, the gap narrows to 2.18 points. The early rule shortfall therefore mainly concerns content not yet correctly recorded, rather than loss of previously credited content. This decomposition does not equate the difficulty or evidence exposure of structural and rule claims.

\subsection{Environmental Change and Experience Maintenance}
\label{app:change-maintenance}
\subsubsection{Selective Revision of Previously Correct Experience}
\label{app:analysis-selective}
To distinguish revision from first acquisition, we track relations credited immediately before a change. A model/unit/event case is eligible when every changed target has a credited one-to-one predecessor for the same semantic relation and at least one unchanged target is credited. The preserved set contains only these previously credited unchanged targets and remains fixed. This yields 487 cases across 20 models, eight environments, and 11 events.

For a changed or preserved target set $A$, coverage at post-event offset $k$ is
\begin{equation}
  C_{A,k}=100\frac{\sum_{a\in A}w_a c_{a,k}}{\sum_{a\in A}w_a},
  \label{eq:analysis-selective}
\end{equation}
where $c_{a,k}$ is current-target credit and $w_a$ is the reviewed lineage weight, normalized within the target set. We average units within events, events within model/environment pairs, eligible environments within models, and models equally. Membership and weights remain fixed across the four offsets. Case eligibility can differ across models because it depends on prior acquisition. Many-to-many replacements and partitioned targets, including Railway pool replacements and Cloud partitioning, lack one-to-one predecessors and are excluded, not scored as failures.

Updated-content coverage rises from 21.47\% to 43.07\%, 59.40\%, and 64.63\% over the first four affected tasks. Preserved-content coverage is 89.05\%, 88.07\%, 90.38\%, and 91.07\%. At the fourth task, preservation exceeds updating in 19 of 20 model means and all eight environment means. Revision thus lags retention in this cohort. These trajectories follow the actual task schedule, and changed and preserved targets can differ in complexity and when they become relevant or observable.

To examine revision after replacement evidence becomes visible, we inspect the Filesystem subset: 122 changed model--relation pairs in 70 eligible model/unit/event cases across 20 models. We match each replacement binding to a successful \texttt{list\_delivery\_lanes} response by workspace, lane, and destination, then check its existing post-task claim score. Exposure accumulates within T07--T10 before each checkpoint. Table~\ref{tab:revision-evidence} partitions this fixed cohort using raw counts, unlike the macro-averaged coverage above. ``No mapped read'' does not exclude evidence from other channels, such as history or successful writes.

\begin{table}[!t]
  \centering
  \small
  \setlength{\tabcolsep}{8pt}
  \caption{Replacement evidence and saved-content states in Filesystem. Each column partitions the same 122 model--relation pairs.}
  \label{tab:revision-evidence}
  \begin{tabular}{@{}lrrrr@{}}
    \toprule
    Revised-content state & T07 & T08 & T09 & T10 \\
    \midrule
    Currently credited & 13 & 59 & 91 & 92 \\
    Exposed, never credited & 8 & 9 & 11 & 12 \\
    Exposed, previously credited but now uncredited & 0 & 0 & 6 & 14 \\
    No mapped read & 101 & 54 & 14 & 4 \\
    \bottomrule
  \end{tabular}
\end{table}

Of the 118 pairs exposed by T10, 104 receive credit at the checkpoint of their first matched read. Yet 26 exposed pairs lack credit at T10: 12 have never received it, and 14 received it earlier but lost it. Saved records illustrate two failures. At T07, Qwen3.5-35B-A3B records new Ember lane identifiers in memory while retaining conflicting old identifiers in a skill. Qwen3.8-Max records and receives credit for the new identifiers at T08, but omits them at T10. Observing replacements therefore does not ensure that saved experience stays current and internally consistent.

\begin{figure*}[!t]
  \centering
  \includegraphics[width=\textwidth]{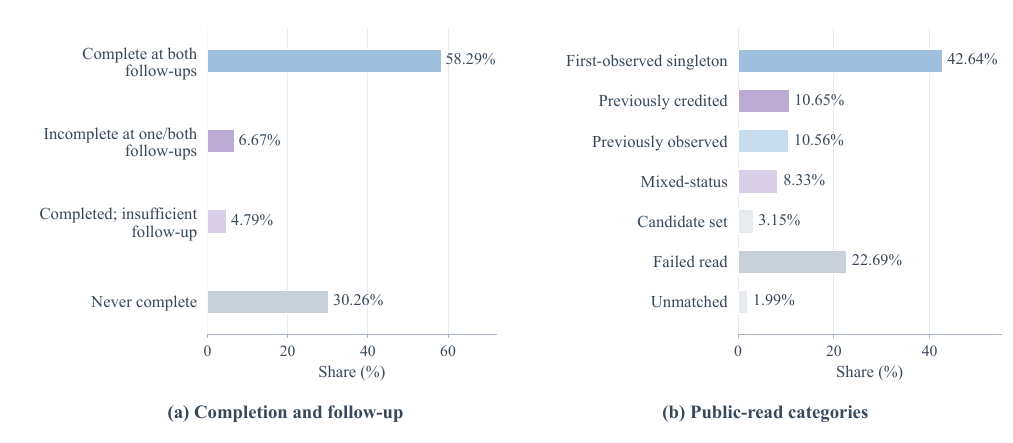}
  \caption{Update persistence and read categories. (a) Completion and two-task follow-up for 2,340 affected model/unit/event observations. (b) Evidence from 2,160 environment queries in Filesystem and Oracle. Bars show within-panel proportions.}
  \label{fig:app-maintenance-reads}
\end{figure*}

\subsubsection{Persistence after First Completion}
\label{app:revision-persistence}
RT records first completion but not persistence. We therefore follow all 2,340 affected model/unit/event observations, including those outside the prior-correct cohort. Completion requires simultaneous credit for all current update targets $N$ and the fixed unchanged targets $V$ credited before the event. With empty $V$, it measures update completion only. We track each first completion over two further checkpoints within the same reference epoch.

Of 1,632 observations that complete at least once, 1,520 have both follow-ups. Among them, 1,364 remain complete at both checkpoints and 156 (10.26\%) become incomplete at one or both. Another 112 completions lack sufficient follow-up, and 708 observations never complete. Figure~\ref{fig:app-maintenance-reads}(a) shows all four outcomes. The 10.26\% regression rate uses only the 1,520 eligible completions. First completion thus does not guarantee completeness over the next two tasks. Regression here means lost credit, not necessarily permanent forgetting or impaired task performance.

\subsection{Interaction and Action Allocation}
\label{app:analysis-actions}

\subsubsection{Tool Use Across Environments}
\label{app:tool-behavior}
Figure~\ref{fig:rq3} uses 151,569 foreground tool calls from all 7,040 with-experience tasks. Tool interfaces determine the categories. Environment tools query or act on the current environment. History tools retrieve prior episodes and records. Experience tools access or modify saved memory and skills. We count explicit requests, including failures, and exclude automatic memory loading and background review. A tool round is an assistant turn containing one or more calls with matching returns. Write counts include recognized requests to create, modify, or delete experience. An unidentified operation remains in the experience-call total but not the write count. One write can affect multiple entries, and the number of writes does not measure the amount or correctness of revised content.

For Figure~\ref{fig:rq3}(a), we average tasks within each environment and then weight the 18 environments equally, retaining zero-call tasks. Panels (b) and (c) use all 26 changes with the same observations at every offset from $-3$ to $+3$. We average events within environments before averaging environments. The overall mean then equally weights all 20 models. In panel (c), Before averages offsets $-3$ to $-1$, and At change denotes the first affected task at offset zero. Task requirements vary across positions, so these comparisons describe observed behavior rather than isolate a causal effect of change. An unchanged write count does not rule out different written content, later writes, or updates during background review.

\subsubsection{Request-Level Action Paths}
\label{app:request-action-paths}
To examine the actions behind request outcomes, we use the Filesystem and Oracle trace cohort described in Appendix~\ref{app:analysis-requests}. Its 8,826 foreground calls comprise 2,189 environment actions, 2,160 environment queries, 1,813 experience operations, and 2,664 history queries. Counts include failures but exclude automatic memory injection and background review.

\begin{figure}[!t]
  \centering
  \includegraphics[width=\textwidth]{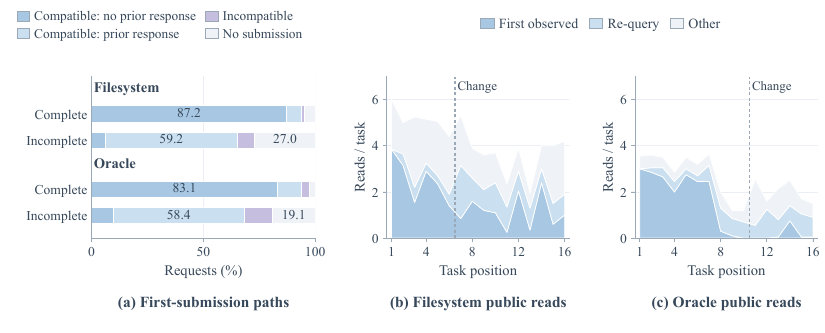}
  \caption{First-submission paths and information gathering in Filesystem and Oracle.}
  \label{fig:app-request-actions}
\end{figure}

First-submission paths use only evidence visible before the assistant turn that generates the arguments. A parallel read in that turn cannot inform them. We classify requests by first-submission compatibility and whether a successful response for their scope preceded submission, retaining no-submission and unknown categories. The matched cohort has no unknown cases. Figure~\ref{fig:app-request-actions}(a) merges the two incompatible-submission paths. A response within a request's scope need not resolve every required binding. Compatible first submissions before successful feedback about the request account for 85.14\% of complete-prior requests and 7.92\% of incomplete-prior requests. This association appears before such feedback could guide a correction, although the recorded sequence does not identify which information source the agent used.

Read categories describe returned evidence, not intent. A \emph{first-observed singleton} supplies one accepted current binding with no earlier singleton or candidate-set exposure through the instrumented interfaces and no same-version prior credit. A \emph{prior-credited query} returns a binding credited at checkpoint $t-1$. A \emph{re-observed singleton} repeats an earlier visible binding when prior credit is absent or unavailable. The latter two categories are \emph{re-queries}. Candidate sets include the accepted value among alternatives, while mixed responses contain multiple statuses. Failed and unmatched reads remain separate. Earlier evidence carries across a change only if its type, unit, condition, field, and complete accepted-value set are unchanged.

Figure~\ref{fig:app-maintenance-reads}(b) counts each environment query once: 42.6\% are first-observed singletons and 21.2\% are re-queries.

The trajectories in Figure~\ref{fig:app-request-actions}(b) and (c) average calls per task across all 20 models, including tasks with no reads. Across complete pre-change and post-change epochs, first-observed singletons fall from 2.53 to 1.14 per task in Filesystem and from 1.86 to 0.15 in Oracle. The epochs are T01--T06 versus T07--T16 in Filesystem and T01--T10 versus T11--T16 in Oracle. Re-queries rise from 0.39 to 1.08 and from 0.47 to 0.84, respectively. These means weight tasks equally, unlike pooled call proportions, and do not isolate the effect of change.

In the matched 1,304-request cohort across 18 models, 201 requests receive no linked submission: 133 in Filesystem and 68 in Oracle. We inspect terminal resource receipts and states to check whether omissions coincide with exhausted budgets. Filesystem capacity is the smaller of the remaining write attempts and remaining calls to environment tools. Oracle capacity is the shared balance divided by the two-unit correction cost, rounded down. All 133 Filesystem omissions retain capacity for a write attempt per omitted request. In Oracle, 61 of 68 omissions leave insufficient budget for even one correction. This contrast also holds with complete prior bindings: all 10 Filesystem omissions retain write capacity, while all six Oracle omissions do not. The eight Filesystem tasks containing those 10 omissions end normally below the model-turn limit.

These balances describe terminal resources, not availability at decision time or delivery under alternative actions. Appendix~\ref{app:oracle-budget-case} illustrates the trade-off between checking and execution at Oracle T10. Call counts exclude tokens, latency, and background work. Domain-specific resource units are not pooled.

% A later Oracle trace shows that successful delivery can coexist with an incorrect explanation of change. At T14, DeepSeek-V4-Pro-0813 retains a skill stating that policy validity ``flips per episode,'' although the four affected cell policies change once at T11 and remain fixed thereafter. Trying two candidate configurations completes all four corrections in six write attempts, including two policy mismatches. This uses 12 of 17 units and scores 100 without new environment queries. The fallback succeeds despite the saved skill's incorrect explanation.
% 
% \clearpage
\section{Illustrative Case Studies}
\label{app:experience-cases}

\subsection{Local Revision and Subsequent Reuse}
\label{app:grove-lifecycle}

\begin{figure}[p]
\centering
\includegraphics[width=\linewidth]{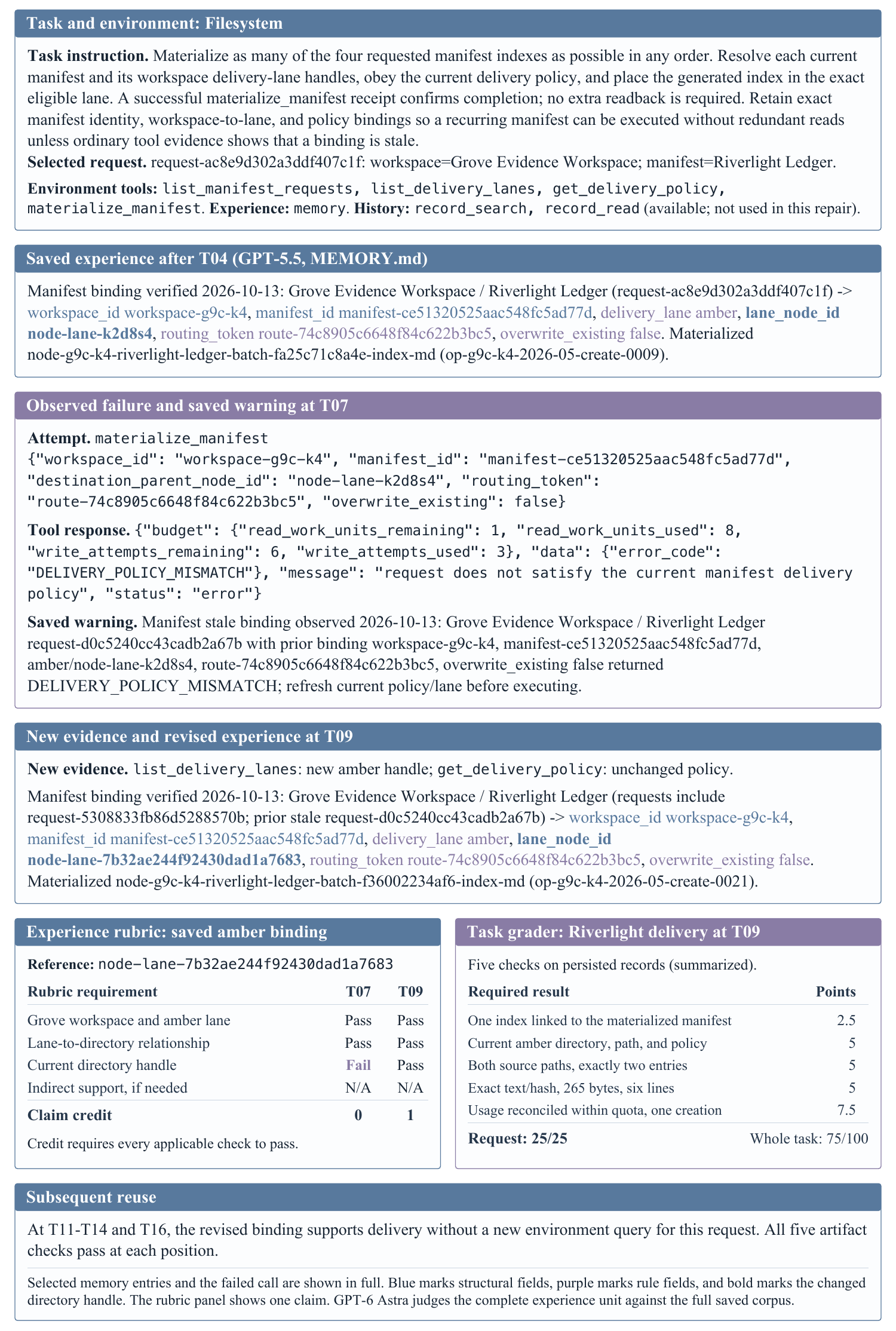}
\caption{Local revision and subsequent reuse in Filesystem. Selected complete records link the revised binding to experience judgments and delivery checks.}
\label{fig:filesystem-case-study}
\label{fig:case-grove-lifecycle}
\end{figure}

Figure~\ref{fig:filesystem-case-study} follows GPT-5.5's Riverlight Ledger binding after a silent destination change before T07. At T09, the agent rechecks public evidence and updates the directory handle while preserving five valid executable fields. The revised binding supports five later deliveries (T11--T14 and T16) without further environment queries for the manifest.

The case also contrasts the two scoring procedures. The experience claim specifies the directory for Grove's amber lane. Applying the reference rubric to saved content, the LLM judge scores it 0 at T07 and 1 at T09. The programmatic task grader instead checks the delivered artifact and final state. Its five checks award the selected T09 request 25/25 points, while the full task scores 75/100.

\clearpage
\subsection{Using Saved Experience under a Budget}
\label{app:oracle-budget-case}

Figure~\ref{fig:oracle-experience-action} illustrates the execution difficulty examined in Appendix~\ref{app:analysis-requests}: complete saved experience may still yield incomplete delivery under a resource budget. At the start of Oracle Clinical One T10, all four models have recorded the required structural and rule bindings. Four corrections cost eight units, leaving room for one five-unit context read within the 17-unit budget. A second read makes full delivery unaffordable. GPT-5.4 spends 15 units on three reads and completes one correction. GPT-5.5 and Claude Opus 5.5 complete all four with eight units. Doubao-Seed-2.1-Pro does so after one read, using 13. GPT-5.4 and Doubao-Seed-2.1-Pro retain an inactivity warning about one requested record, providing context for checking. One targeted check remains compatible with full delivery, while further checks consume the budget needed for execution.

\begin{figure}[!ht]
\centering
\includegraphics[width=\linewidth]{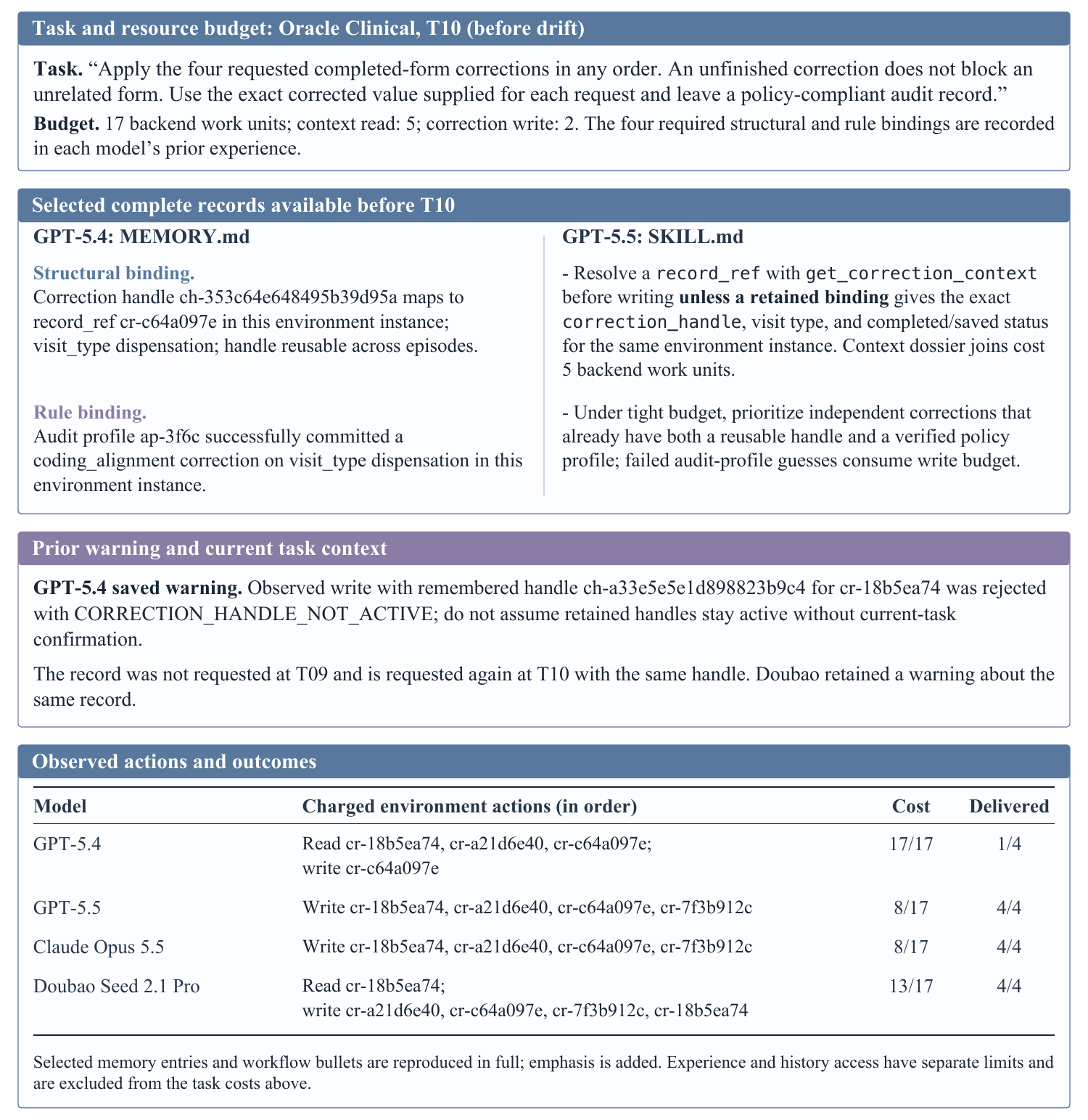}
\caption{Complete saved experience, resource use, and delivery on the same Oracle Clinical One task. Selected entries are reproduced in full.}
\label{fig:oracle-experience-action}
\end{figure}

\end{document}